\documentclass[runningheads]{llncs}
\usepackage{graphicx}

\usepackage{tikz}
\usepackage{comment}
\usepackage{amsmath,amssymb} %
\usepackage{color}
\usepackage{epsfig}
\usepackage{lipsum}
\usepackage{algorithm}
\usepackage{algpseudocode}
\usepackage{url}
\usepackage{xcolor, colortbl}
\usepackage{mathtools}
\usepackage{tabularx}
\usepackage{multirow}
\usepackage{enumitem}
\usepackage{bbm}
\usepackage{wrapfig}
\usepackage{setspace}
\RequirePackage{fix-cm}

\usepackage{graphicx}
\usepackage{amsmath}
\usepackage{amssymb}
\usepackage{booktabs}

\usepackage{color}
\usepackage{amssymb}
\usepackage{lipsum}
\usepackage{algorithm}
\usepackage{algpseudocode}
\usepackage{xcolor}
\usepackage{tabularx}
\usepackage{multirow}
\usepackage{enumitem}
\usepackage{bbm}
\usepackage{wrapfig}
\usepackage{setspace}
\usepackage{footnote}

\usepackage{subcaption}
\usepackage{colortbl} %
\usepackage{pifont}  %
\usepackage{cite}  %
\usepackage{mathtools}  %
\usepackage[font=small,labelfont=bf]{caption}  %
\usepackage{blindtext}  %
\usepackage{stmaryrd} %
\usepackage[accsupp]{axessibility}  %

\usepackage[breaklinks=true,colorlinks,bookmarks=true]{hyperref}

\newcommand{\Fig}[1]{Fig.~\ref{fig:#1}}
\newcommand{\Sec}[1]{Sec.~\ref{sec:#1}}
\newcommand{\Eq}[1]{Eq.~(\ref{eq:#1})}
\newcommand{\Tbl}[1]{Table~\ref{tab:#1}}
\newcommand{\Alg}[1]{Algorithm~\ref{algo:#1}}

\def\ie{\emph{i.e.}}
\def\eg{\emph{e.g.}}

\def\etal{\emph{et al.}}

\definecolor{purp}{rgb}{0.65, 0.16, 0.65}
\definecolor{bblue}{rgb}{0.2, 0.2, 0.6}
\definecolor{brown}{rgb}{0.65, 0.16, 0.16}
\definecolor{orange}{rgb}{1.0, 0.5, 0.0}
\definecolor{blue}{rgb}{0.0, 0.5, 1.0}
\definecolor{green}{rgb}{0, 0.8, 0}
\definecolor{lgreen}{rgb}{0.6, 0.8, 0}
\definecolor{red}{rgb}{0.8, 0, 0}
\definecolor{darkblue}{rgb}{0, 0.2, 0.6}
\definecolor{brinkpink}{rgb}{0.98, 0.38, 0.5}
\definecolor{Gray}{gray}{0.1}

\newcolumntype{C}[1]{>{\centering\let\newline\\\arraybackslash\hspace{0pt}}p{#1}}

\def\etal{\emph{et al.}}

\newcommand{\cmark}{\ding{51}}
\newcommand{\xmark}{\ding{55}}

\definecolor{grey}{rgb}{0.9, 0.9, 0.9}

\DeclareMathOperator*{\argmax}{argmax} %

\newcommand{\bx}{\mathbf{x}}

\newcommand{\bW}{\mathbf{W}}
\newcommand{\bw}{\mathbf{w}}
\newcommand{\bz}{\mathbf{z}}

\newcommand{\textS}{{\text{S}}}
\newcommand{\textL}{{\text{L}}}
\newcommand{\textT}{{\text{T}}}
\newcommand{\textP}{{\text{P}}}
\newcommand{\textPL}{{\text{PL}}}

\newcommand{\veryshortarrow}[1][3pt]{\mathrel{%
   \hbox{\rule[\dimexpr\fontdimen22\textfont2-.2pt\relax]{#1}{.4pt}}%
   \mkern-4mu\hbox{\usefont{U}{lasy}{m}{n}\symbol{41}}}}
  
\newcommand{\hyperfootnote}[1][]{\def\ArgI\hyperfootnoteRelay}
\newcommand\hyperfootnoteRelay[2][]{\href{#1#2}{\ArgI}\footnote{\href{#1#2}{#2}}}

\begin{document}
\pagestyle{headings}
\mainmatter
\def\ECCVSubNumber{2012}  %

\title{Combating Label Distribution Shift for\\Active Domain Adaptation}

\titlerunning{Combating Label Distribution Shift for Active Domain Adaptation}
\pdfstringdefDisableCommands{%
  \def\\{}%
}

\author{Sehyun Hwang\inst{1} \qquad
Sohyun Lee\inst{2}
\qquad
Sungyeon Kim\inst{1} 
\\
Jungseul Ok\inst{1,2}\thanks{Co-corresponding authors} 
\qquad
Suha~Kwak\inst{1,2}$^\star$
}
\authorrunning{Sehyun Hwang, Sohyun Lee, Sungyeon Kim, Jungseul Ok, and Suha Kwak}
\institute{
$^{1}$Department of Computer Science and Engineering, POSTECH, Korea\\
$^{2}$Graduate School of Artificial Intelligence, POSTECH, Korea\\
{\tt\small \url{http://cvlab.postech.ac.kr/research/LAMDA/}}
}
\maketitle

\begin{abstract}
We consider the problem of active domain adaptation (ADA) to unlabeled target data, of which subset is actively selected and labeled given a budget constraint.
Inspired by recent analysis on a critical issue from label distribution mismatch between source and target in domain adaptation, we devise a method that addresses the issue for the first time in ADA.
At its heart lies a novel sampling strategy, which seeks target data that best approximate the entire target distribution as well as being representative, diverse, and uncertain. 
The sampled target data are then used not only for supervised learning but also for matching label distributions of source and target domains, leading to remarkable performance improvement. 
On four public benchmarks, our method substantially outperforms existing methods in every adaptation scenario.

\end{abstract}

\section{Introduction} \label{sec:intro}
Domain adaptation is the task of adapting a model trained on a label-sufficient source domain to a label-scarce target domain when their input distributions are different.
It has played crucial roles in applications that involve significant input distribution shifts such as recognition under adverse conditions (\eg, climate changes~\cite{dai2020curriculum, Sakaridis_2018_IJCV, Sakaridis_2018_ECCV, lee2022fifo} and nighttime~\cite{sakaridis2019guided}) and synthetic-to-real adaptation~\cite{peng2018visda}.
The most popular direction in this field is unsupervised domain adaptation~\cite{ben2010theory, dann} which assumes a totally unlabeled target domain.
However, in practice, 
labeling a small part of target data is usually feasible.
Hence, label-efficient domain adaptation tasks such as semi-supervised domain adaptation~\cite{mme, decota, cdac, ecacl, yoon2022semi} and active domain adaptation~\cite{su2020active, tqs, clue, s3vaada} have attracted increasing attention.

In this paper, we consider 
active domain adaptation (ADA) \cite{su2020active, tqs, clue, s3vaada},
where
we can interact with an oracle
to obtain annotations
on a subset of target data given budget constraint, while utilizing the annotations for domain adaptation.
The key to the success of ADA is to 
co-design sampling mechanism 
selecting a subset of target data to be annotated and utilization of the annotations.
Existing ADA methods utilize the obtained annotations only for supervised learning, similar to existing Active Learning (AL) methods~\cite{sener2017active, sinha2019variational, ash2019deep, wang2019incorporating}.
Accordingly, they count diversity, representativeness, and uncertainty of the data to boost the effect of supervised learning.

We argue that for domain adaptation, there is another use of the sampled data, which deserves attention but is missing in the previous work: matching label distributions of source and target domains. 
In practice, domain adaptation
often encounters
label distribution shift, i.e., the frequencies of classes significantly differ between source and target domains.
It has been proven in~\cite{zhao19, tachet2020domain} that 
matching label distributions of source and target domains is a necessary condition for successful domain adaptation~\cite{zhao19, tachet2020domain}.
Also, it has been empirically verified in~\cite{tachet2020domain} that mismatched label distributions restrict or even deteriorate performance of existing domain adaptation methods~\cite{dann,cdan,long2017deep}.

Motivated by this, we present a new method that addresses the label distribution shift for the first time in ADA.
At the heart of our method lies LAbel distribution Matching through Density-aware Active sampling, and thus it is dubbed LAMDA.
Its key idea is to use sampled data for label distribution matching as well as supervised learning.
During training, it estimates the label distribution of the target domain through the annotated labels of sampled target data, and 
builds each source data mini-batch in a way that the label frequencies of the batch follow the estimated target label distribution.
To this end, we design a new sampling strategy useful for label distribution estimation as well as supervised learning.
For supervised learning, sampled data are encouraged to be representative, diverse, and uncertain.
For label distribution estimation, on the other hand, sampled data should well approximate the entire data distribution of the target domain.
As will be demonstrated empirically, existing ADA methods often fail to satisfy the second condition since they blindly select uncertain instances or do not take the overall target distribution into account.

\begin{figure}[!t]
\begin{center}
\includegraphics[width= \linewidth]{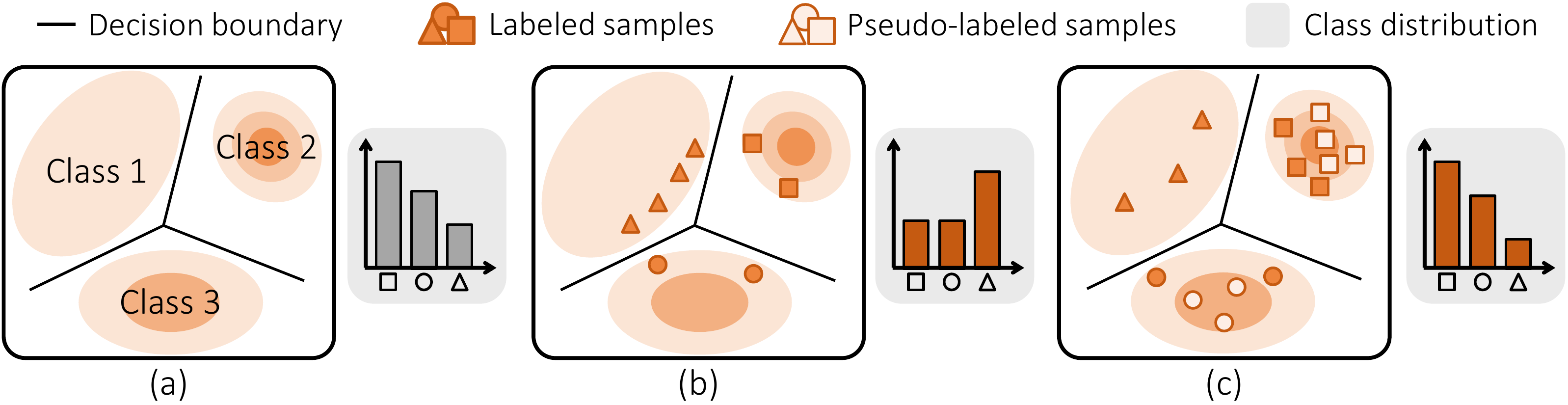}
\end{center}
\vspace{-4mm}
\caption{
Comparison between sampling methods.
(a) Data distribution and label distribution of target data.
(b) Uncertainty preferred sampling of conventional ADA and the label distribution of corresponding sampled data.
(c) Density-aware sampling of LAMDA and the label distribution of corresponding sampled data.
}
\label{fig:sampling}
\vspace{-3.5mm}
\end{figure}

Our sampling method satisfies both of the above conditions.
Specifically, it selects a subset of target data whose statistical distance from the entire target data is minimized.
Since the distribution of the sampled data well approximates that of the entire target data, their labels are expected to follow the latent target label distribution.
They also spontaneously become diverse and representative in order to cover the entire target data distribution.
In addition, LAMDA asks the oracle for labeling only uncertain instances in the sampled subset;
it in turn utilizes the manually labeled samples for both supervised learning and label distribution estimation, while the rest are assigned pseudo labels by the model's prediction and used only for label distribution estimation.
This strategy lets LAMDA annotate and exploit only uncertain data in the subset for supervised learning, and estimate the target label distribution accurately by using the entire subset.
The advantage of our sampling method is illustrated in~\Fig{sampling}. %

In addition, we propose to use the cosine classifier~\cite{gidaris2018dynamic, qi2018low}, instead of the conventional linear classifier, in order to further alleviate the adverse effect of label distribution shift.
The cosine classifier is known to be less biased to dominant classes %
since its classification weights are $\ell_2$-normalized, and thus has been used for long-tailed recognition~\cite{Kang2020Decoupling} and few-shot learning~\cite{qi2018low, gidaris2018dynamic, chen2018a}.
We find that such a property is also useful to combat label distribution shift; it is empirically verified that the cosine classifier significantly improves ADA performance when combined with a domain alignment method.

To evaluate and compare LAMDA with existing ADA methods thoroughly, we present a unified evaluation protocol for ADA.
Extensive experiments based on the evaluation protocol demonstrate impressive performance of LAMDA, which largely surpasses records of existing ADA methods~\cite{tqs, clue, s3vaada}, on four public benchmarks for domain adaptation~\cite{officehome, tan2020class, domainnet, peng2018visda}.
The main contribution of this paper is four-fold:
\begin{itemize}[leftmargin=5mm]
\item LAMDA is the first attempt to tackle the label distribution shift for ADA. The importance of this research direction is demonstrated by the outstanding performance of LAMDA.
\item We propose a new sampling strategy for choosing target data best preserving the entire target data distribution as well as being representative, diverse, and uncertain.
Data selected by our strategy are useful for both label distribution matching and supervised learning.
\item For the first time, we benchmark existing ADA methods~\cite{tqs, clue, s3vaada} on four public datasets for domain adaptation~\cite{officehome, tan2020class, domainnet, peng2018visda} through a unified evaluation protocol.
\item In our experiment with each of the four domain adaptation datasets,
LAMDA substantially outperforms all the existing ADA models.
\end{itemize}
\section{Related work} \label{sec:related}
\noindent \textbf{Unsupervised domain adaptation (UDA).}
Major approaches in UDA aim at learning domain invariant features so that a classifier trained on the labeled source domain data can be transferred to the unlabeled target domain data~\cite{ben2010theory}.
To do so, previous methods align feature distribution between the two domains using various domain discrepancy measures such as
MMD~\cite{mkmmd, long2017deep}, Wasserstein discrepancy~\cite{courty2017joint, courty2016optimal, damodaran2018deepjdot, lee2019sliced}, and $\mathcal{H}$-divergence~\cite{ajakan2014domain, dann, cdan, gvb, purushotham2016variational, tzeng2015simultaneous}.
On the other hand, recent studies~\cite{zhao19, tachet2020domain} found that such domain alignment is only effective when the label distributions of the two domains are matched.
This condition is difficult to be satisfied due to the limited access to the target class distribution.
In this work, we propose to utilize the actively sampled data in ADA to estimate the target label distribution and match the label distribution of the two domains for the effective domain alignment.

\noindent \textbf{Active learning (AL).}
AL is a task of selecting the most performance-profitable samples to be annotated from an oracle~\cite{settles2009active}.
Previous methods design various selection strategies, where they often refer to uncertainty~\cite{asghar2016deep, he2019towards, ostapuk2019activelink}, diversity~\cite{sener2017active, sinha2019variational}, or the both~\cite{ash2019deep, wang2015querying, wang2019incorporating} for the selection.
Uncertainty-based methods prefer difficult samples for the model, \eg, samples with high entropy.
Diversity-based methods prefer samples that are different from the selected ones.
Our method shares a similar idea with Wang~\cite{wang2015querying, wang2019incorporating} in that we use MMD~\cite{mmd}, but we additionally select the easy-but-representative samples as a pseudo-labeled set to precisely estimate the target label distribution, which can be used to help domain adaptation process.

\noindent \textbf{Active domain adaptation (ADA).}
ADA is a variant of active learning that selects samples to maximize the domain adaptation performance.
ADA is first introduced by Rai~\etal~\cite{rai2010domain} and first adapted to image classification by AADA~\cite{su2020active}.
Existing methods mainly refer to the difficulty of samples (\ie, uncertainty) for selection.
TQS~\cite{tqs} selects uncertain samples by combining three sampling criteria: disagreement among ensemble models~\cite{seung1992query}, top-2 margin of predictive probabilities~\cite{roth2006margin}, and confidence of domain discriminator~\cite{su2020active}.
CLUE~\cite{clue} additionally considers the diversity among the selected samples along with the uncertainty by using entropy-weighted $k$-means clustering~\cite{huang2005automated}.
$\text{S}^\text{3}$VAADA~\cite{s3vaada} designs a set-based scoring function that favors three properties: vulnerability to adversarial perturbation, diversity within the sampled set, and representativeness to avoid outliers.
More recent methods utilize a free energy biases~\cite{lecun2006tutorial} of the two domains~\cite{xie2022active}, K-medoids algorithm~\cite{rdusseeun1987clustering, deheeger2021discrepancy}, and the distance to different class centers~\cite{xie2022learning} for the selection.
To newly tackle the critical issue of label distribution shift in ADA, we propose a sampling strategy that considers the data distribution of the target domain.
The main technical difference between the sampling of conventional ADA and ours is illustrated in~\Fig{sampling}.

\section{Problem formulation} \label{sec:problem}
Given a labeled source dataset $\mathcal{D}_\textS = \{(\bx_i, y_i)\}_{i=1}^{n_\textS}$ of size $n_\textS$ and an unlabeled target dataset $\mathcal{D}_\textT=\{\bx_i\}_{i=1}^{n_\textT}$ of size $n_\textT$,
we study a standard ADA scenario of 
$R$~rounds, in each of which
$B$ samples of target data are newly labeled and utilized for model update, \ie,
the per-round budget is $B$ and the total budget is $RB \le n_\textT$.
Let $\mathcal{D}_\textL$ be the labeled target dataset actively collected, which
grows up to size $RB$.
We consider image classification such that 
$\bx_i$ is an image and $y_i \in \mathcal{Y} = \{1,2, \ldots, C\}$ is a categorical variable,
where a model, parameterized by $\boldsymbol{\theta}$, predicts  $\argmax_{y \in \mathcal{Y}} p_{\boldsymbol{\theta}}(y|\bx)$ for input image $\bx$.
The goal of ADA is to maximize the test accuracy of $\boldsymbol{\theta}$ 
in the target domain, where $\boldsymbol{\theta}$ is trained on $\mathcal{D}_\textS$, $\mathcal{D}_\textT$ and $\mathcal{D}_\textL$ in the iterative manner.

\section{Proposed method} \label{sec:method}
\subsection{Overview of LAMDA} \label{sec:overall}

\begin{figure*}[t]
    \centering
    \includegraphics[width=\linewidth]{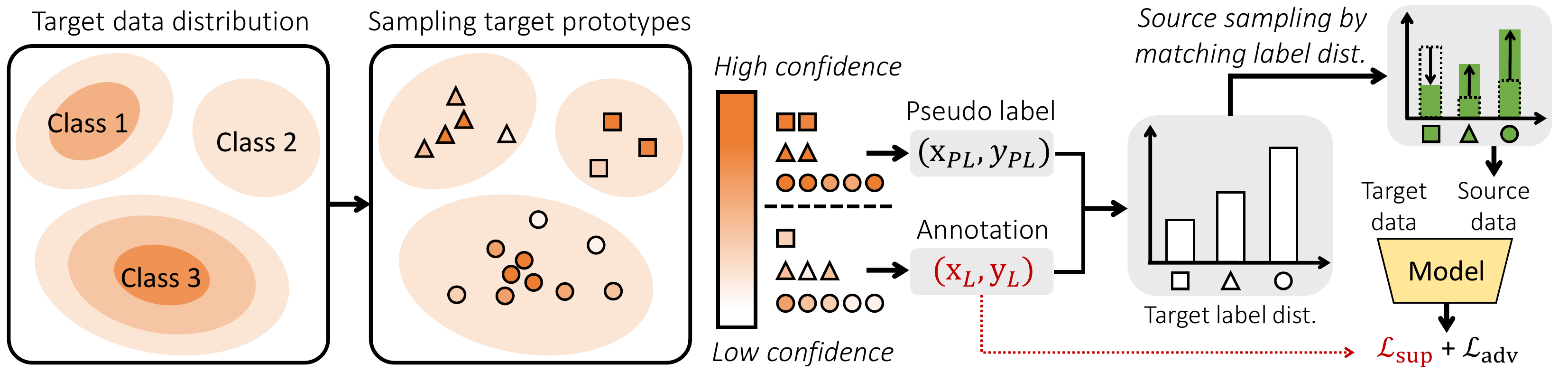}
\vspace{-5mm}
\caption{
LAMDA first samples a set of target prototypes that well represent the entire target data distribution.
The prototypes are annotated by an oracle if their predictions are uncertain or assigned pseudo labels otherwise.
It then estimates the target label distribution using the prototypes, and builds source data mini-batches whose label distributions follow the estimated target label distribution.
Finally, our model is trained by the cross-entropy loss $\mathcal{L}_\text{sup}$ and the domain adversarial loss $\mathcal{L}_\text{adv}$.
} \label{fig:pipeline}
\vspace{-2mm}
\end{figure*}

We present a novel ADA method, named LAMDA, that addresses \emph{label distribution shift} between source and target domains. 
Our core idea is to select and utilize target samples useful for both label distribution matching and supervised learning.
This idea is implemented in LAMDA by three components: prototype sampling, label distribution matching, and model training.
First, LAMDA selects a set of prototypes, \ie, target data that best approximate the entire target data distribution.
Uncertain prototypes in the set are then identified by the model and annotated by oracle, while the rest are assigned pseudo-labels (\Sec{prototype}).
Next, LAMDA estimates the target label distribution using the assigned labels of the prototypes and adjusts the label distribution of source data being drawn within each mini-batch according to the estimated target label distribution (\Sec{calibrated_dann}).
Under the matched label distribution, the model is trained by both cross-entropy loss and domain adversarial loss (\Sec{training}).
The overall framework of LAMDA is illustrated in~\Fig{pipeline}.
In what follows, we describe each component at a round.

\subsection{Prototype set sampling in target data} \label{sec:prototype}

We begin with a model $\boldsymbol{\theta}$
which is 
from the previous round, or pretrained on source dataset $\mathcal{D}_\textS$ for the first round.
Let $X_{(\cdot)}$ denote the 
set of images in dataset $\mathcal{D}_{(\cdot)}$ for notational simplicity.
To select the prototype set that represents the target data distribution, 
we first seek subset $X \subset X_\textT$ which minimizes a statistical distance between $X$ and the entire target data $X_\textT$.
Inspired by the sampling technique for example-based model explanation~\cite{mmdcritic},
we employ
the squared Maximum Mean Discrepancy (MMD)~\cite{mmd} between $X$ and $X_\textT$ on the feature space, which is formally given by
\begin{equation}
\begin{aligned}
    \text{MMD}^2(X,X_\textT) :=\,&\frac{1}{|X|^2}\sum_{\bx_i,\bx_j \in X} k(f(\bx_i),f(\bx_j))
    \\
    &- \frac{2}{n_\textT|X|}\sum_{\bx_i\in X, \bx_j\in X_\textT} k(f(\bx_i),f(\bx_j))\\
                         &+ \frac{1}{n_\textT^2}\sum_{\bx_i,\bx_j\in X_\textT} k(f(\bx_i),f(\bx_j)) \;,
    \label{eq:mmd2}    
\end{aligned}
\end{equation}
where we let $f(\bx)$ be the feature of input $\bx$ extracted by $\boldsymbol{\theta}$,
and $k(\bz, \bz') = \text{exp}(-\gamma||\bz - \bz'||^2)$ be the Radial Basis Function (RBF) kernel.
Noting that the last term in \Eq{mmd2} is constant with respect to $X$,
we define $J(X)$ as follows:
\begin{equation}
\begin{aligned}
    J(X) := &\,\text{MMD}^2(\emptyset,X_\textT) - \text{MMD}^2(X,X_\textT)\\
         = &\,\frac{2}{n_\textT|X|}\sum_{\bx_i\in X,\bx_j\in X_\textT} k(f(\bx_i),f(\bx_j))
           - \frac{1}{|X|^2}\sum_{\bx_i,\bx_j \in X} k(f(\bx_i),f(\bx_j)) \;.
    \label{eq:cost}
\end{aligned}
\end{equation}
where a constant $\text{MMD}^2(\emptyset,X_\textT)$ is added to make $J(\emptyset) = 0$, and the first and second terms measure representativeness and diversity of $X$, respectively.

The prototypes can be then identified by a constrained combinatorial optimization
to maximize $J(X)$ %
given a certain size limit $n_\textP$, \ie,
\begin{equation}
    \max_{X \in 2^{X_\textT}: |X| \leq n_\textP} J(X) \;.
     \label{eq:objective}
\end{equation}
This is generally intractable due to the exponentially many candidates.
However, a greedy process selecting samples one after one to locally maximize $J$
can efficiently find a near-optimal solution in polynomial time
since $J(X)$ is monotone submodular when $k$ is RBF kernel~\cite{mmdcritic}.
To be specific, 
the greedy process is proven to achieve at least $1-[(n_\textP-1)/n_\textP]^{n_\textP}$ of the optimum~\cite{nemhauser1978analysis}.
We hence adopt the greedy process to select subset $X_\textP$ from the unlabeled target data.

We note that
setting $n_\textP = B$ and spending all the budget for $X_\textP$ 
would be a waste of budget when $X_\textP$ includes easy prototypes, whose labels 
are accurately predicted by $\boldsymbol{\theta}$.
We hence set $n_\textP$ in an adaptive way so that we spend budget $B$ only for hard prototypes. To be specific,
starting from 
$X_\textPL = \emptyset$ and $X_\textL$ from the previous round
(or $X_\textPL = X_\textL = \emptyset$ for the first round),
each greedy selection is added to either $X_\textPL$ or $X_\textL$.
$X_\textPL$ includes only easy prototypes of $X_\textP$ whose margin between top-1 and top-2 predictions is larger than threshold $\Delta$, and
only hard prototypes in $X_\textL = X_\textP \setminus X_\textPL$ are labeled by oracle.
For $X_\textPL$, we use top-1 prediction as the pseudo label which is given by
\begin{equation}
    \hat{y}_{i,1} := \argmax_{y\in\mathcal{Y}} p_{\boldsymbol{\theta}}(y|\bx_i) \;.
    \label{eq:pseudo_lbl}
\end{equation}
In each round,
we continue the sampling process until $B$ hard samples are newly annotated by oracle. Thus, $n_\textP = |X_\textP| \ge B$ is determined by the adaptation scenario and the model in hand.
This is possible because the greedy selection can return a near-optimal solution at any iteration.
The sampling process is illustrated in Fig.~\ref{fig:pipeline}, and described formally in~\Alg{sampling}.

We denote the set of labeled prototypes by $\mathcal{D}_\textL = \{(\bx_i, y_i)\}_{i=1}^{B}$ and that of pseudo-labeled prototypes by $\mathcal{D}_\textPL = \{(\bx_i, \hat{y}_{i,1})\}_{i=1}^{n_\textPL}$.
$\mathcal{D}_\textL$ is used for both supervised learning and label distribution estimation,
while $\mathcal{D}_\textPL$ is used only for label distribution estimation;
details will be described in the following section.

\begin{algorithm}[t!]
\fontsize{8}{10}\selectfont
\caption{Prototype sampling at a round}
\label{algo:sampling}
\renewcommand{\algorithmicensure}{\textbf{Define:}}
\begin{algorithmic}[1]
\Require Model parameter $\boldsymbol{\theta}$ from the previous round, labeled source dataset $\mathcal{D}_\textS$, unlabeled target image set $X_\textT$, per-round budget $B$, threshold $\Delta$.
\State Retrieve $X_\textL$ from the previous round or set it as empty set for the first round.
\State Set $X_\textPL \leftarrow \emptyset$ and $X_\textP \leftarrow X_\textL \cup X_\textPL$.
\Repeat
    \State $\bx^* \gets \argmax\limits_{\bx_i \in X_\textT \setminus X_\textP}(J(X_\textP \cup \{\bx_i\}) - J(X_\textP))$
    \Comment{Prototype selection w.r.t. $J(X)$ in \Eq{cost}}
    \State $\hat{y}_{1} \gets \argmax\limits_{y\in\mathcal{Y}} p_{\boldsymbol{\theta}}(y|\bx^*)$,\; $\hat{y}_{2} \gets \argmax\limits_{{y\in\mathcal{Y}} \setminus \{\hat{y}_{1}\}} p_{\boldsymbol{\theta}}(y|\bx^*)$   
    \Comment{Get top-1 and top-2 prediction}
    \If{$p_{\boldsymbol{\theta}}(\hat{y}_{1}|\bx^*) - p_{\boldsymbol{\theta}}(\hat{y}_{2}|\bx^*) > \Delta$}
        \Comment{Identify easy/hard prototype by margin}
        \State $X_\textPL \gets X_\textPL \cup \{\bx^*\}$
        \Comment{Pseudo-labeling for easy prototype}
    \Else
        \State $X_\textL \gets X_\textL \cup \{\bx^*\}$
        \Comment{Oracle-labeling for hard prototype}
    \EndIf
\State $X_\textP \gets X_\textP \cup \{\bx^*\}$
\Comment{$X_\textP = X_\textL \cup X_\textPL$}
\Until{$B$ samples are newly added to $X_\textL$ (and labeled by oracle)}
\State \Return  $X_\textP, X_\textPL, X_\textL$
\end{algorithmic}
\end{algorithm}

\subsection{Label distribution matching} \label{sec:calibrated_dann}
We use the prototype set to estimate the target data distribution $p_\textT(y)$, which is in turn used for label distribution matching.
To estimate $p_\textT(y)$, we investigate the frequency of each class within $\mathcal{D}_\textL$ and $\mathcal{D}_\textPL$.
The frequency of class $c$ in $\mathcal{D}_\textL$ is computed by
\begin{equation}
    n_{\textL,c} := \sum_{(\bx_i, y_i) \in \mathcal{D}_\textL} \mathbbm{1}[y_i = c] \;,
    \label{eq:nl}
\end{equation}
where $\mathbbm{1}$ is an indicator function.
On the other hand, the class frequency in $\mathcal{D}_\textPL$ is weighted by the corresponding predictive probability, which is given by
\begin{equation}
    \hat{n}_{\textPL,c} := \sum_{(\bx_i, \hat{y}_{i,1}) \in \mathcal{D}_\textPL} \mathbbm{1}{[\hat{y}_{i,1} = c]}  p_{\boldsymbol{\theta}}(\hat{y}_{i,1}|\bx_i) \;,
    \label{eq:npl}
\end{equation}
Then, the target label distribution $p_\textT(y)$ is estimated by
\begin{equation}
\hat{p}_\textT(y) := 
    \frac{n_{\textL,y} + \hat{n}_{\textPL,y} + 1}{n_\textL + \hat{n}_\textPL + C} \;,
    \label{eq:tgt_lbl}
\end{equation}
where $\hat{n}_\textPL = \sum_c \hat{n}_{\textPL,c}$ and $C$ is the number of classes.
Note that we add an offset 1 to each category frequency of~\Eq{tgt_lbl} to ensure at least a single instance is considered to be present in the target domain.
This is consistent with the assumption of UDA,
where both domains have the same label space $\mathcal{Y}$.
To make the observed source label distribution follow $\hat{p}_\textT(y)$, we apply class-weighted sampling when building source mini-batches.
The ratio between the source label distribution $p_\textS(y)$ and the estimated target label distribution $\hat{p}_\textT(y)$ is denoted by $w(y) := \frac{\hat{p}_\textT(y)}{p_\textS(y)} \;$.
Then, the probability of sampling $(\bx_i, y_i)$ from $\mathcal{D}_\textS$ for source mini-batch construction is defined by
\begin{equation}
    \rho_i :=  \frac{w(y_i)}{\sum_{(\bx_j, y_j) \in \mathcal{D}_\textS} w(y_j)} \;,
    \label{eq:src_sampling}
\end{equation}
where $i$ indicates the sample index.

\subsection{Model training} \label{sec:training}
\noindent\textbf{Loss functions.}
As the label frequencies of a source mini-batch match those of the target domain by~\Eq{src_sampling}, we can now apply a domain alignment loss while alleviating the label distribution shift.
We choose the domain adversarial loss~\cite{dann}, but any other losses~\cite{dann, cdan, long2017deep} for domain alignment can be employed.
For domain adversarial training, a domain discriminator, parameterized by $\boldsymbol{\phi}$, is trained to classify the domain of input feature by probability $p_{\boldsymbol{\theta}_f,\boldsymbol{\phi}}(d|\bx)$, where $d \in \{0,1\}$ is domain label.
In the meantime, the feature extractor parameterized by $\boldsymbol{\theta}_f$ is adversarially trained to confuse the discriminator.
The domain adversarial loss with the matched label distributions is given by
\begin{equation}
    \mathcal{L}_\text{adv} := \,\mathbb{E}_{\bx_i \stackrel{\rho_i}{\sim} X_\textS} [-\log p_{\boldsymbol{\theta}_f,\boldsymbol{\phi}}(d|\bx_i)]
                   + \mathbb{E}_{\bx_j \stackrel{\text{iid}}{\sim} X_\textT} [-\log (1 - p_{\boldsymbol{\theta}_f,\boldsymbol{\phi}}(d|\bx_j))] \;,
    \label{eq:dann}
\end{equation}
where the first expectation is taken over $\rho_i$ of $X_\textS$ and the second one is taken over uniform distribution of $X_\textT$.
The $\boldsymbol{\theta}_f$ is updated to maximize $\mathcal{L}_\text{adv}$, while $\boldsymbol{\phi}$ is updated to minimize $\mathcal{L}_\text{adv}$.
Meanwhile, the cross-entropy loss for labeled data $\mathcal{D}_\textS$ and $\mathcal{D}_\textL$ is given by
\begin{equation}
\begin{aligned}
    \mathcal{L}_\text{sup} := \,\mathbb{E}_{(\bx_i, y_i) \stackrel{\rho_i}{\sim} \mathcal{D}_\textS} [-\log p_{\theta}(y_i|\bx_i)]
                   + \mathbb{E}_{(\bx_j, y_j) \stackrel{\text{iid}}{\sim} \mathcal{D}_\textL} [-\log p_{\theta}(y_j|\bx_j)] \;.
    \label{eq:lbl}
\end{aligned}
\end{equation}
In summary, the total training loss for the proposed framework is given by
\begin{equation}
    \mathcal{L} := \mathcal{L}_\text{sup} + \mathcal{L}_\text{adv} \;.
    \label{eq:loss_all}
\end{equation}

\noindent\textbf{Cosine classifier.} \label{sec:cosine}
To further alleviate the negative effect of label distribution shift, LAMDA employs a cosine classifier~\cite{gidaris2018dynamic, qi2018low}, which measures cosine similarities between the classifier weights and an embedding vector as classification scores. %
The norm of classifier weight is known to be greatly affected by the label distribution~\cite{Kang2020Decoupling, gidaris2018dynamic, xu2019larger}.
Since the norm does not interfere with the classification score in the cosine classifier, it can alleviate the label distribution shift.
Specifically, let $\bW := \{\bw_c\} \in \mathbb{R}^{d \times C}$, where $\bw_c \in \mathbb{R}^{d}$ indicates a weight of classifier for class $c$ with embedding dimension $d$.
Then, the class probability predicted by the cosine classifier is given by
\begin{equation}
    p_{\boldsymbol{\theta}}(y=c|\bx_i) := \textrm{softmax}\bigg(\frac{{h\circ f(\bx_i)}^\top \bw_c}{\tau\,||{h\circ f(\bx_i)}||\,||\bw_c||}\bigg) \;,
    \label{eq:predicted_probability}
\end{equation}
where $h$ is a single hidden layer that projects feature vector $f(\bx)$ into $d$-dimensional embedding space, and $\tau$ is a temperature term that adjusts sharpness of the predicted probability.

\section{Experiments}

We first describe datasets, experiment setup, and implementation details in \Sec{setup}.
Then LAMDA is evaluated and compared with previous work in \Sec{result}, and contributions of its components are scrupulously analyzed in \Sec{analysis}.

\subsection{Setup} \label{sec:setup}

\noindent\textbf{Datasets.}
We use four domain adaptation datasets with different characteristics:
OfficeHome \cite{officehome},
OfficeHome-RSUT \cite{tan2020class},
VisDA-2017 \cite{peng2018visda}, and DomainNet \cite{domainnet}.
OfficeHome %
contains 16k images from four domains \{Art, Clipart, Product, Real\}, where we conduct a diverse set of domain adaptation for each of 12 source-target permutations.
OfficeHome-RSUT %
is a dataset 
sub-sampled from 
three domains \{Clipart, Product, Real\} of OfficeHome,
where
the subsampling protocol, called 
reversely-unbalanced source and unbalanced target (RSUT),
is employed to make a large label shift between source and target domains.
VisDA-2017 %
is a large-scale dataset consisting of 207k images from two domains \{Synthetic, Real\} in a realistic scenario of synthetic-to-real domain adaptation.
DomainNet %
is also a large-scale dataset 
but has a prevalent labeling noise.
In DomainNet, we use five domains \{Real, Clipart, Painting, Sketch, Quickdraw\}\footnote{
The domains are chosen considering their consistency with existing benchmarks~\cite{clue}.}
consisting of 362k images.
We use 10\% of the datasets for validation and the rest are kept for training.
While DomainNet includes an independent test set, the other datasets do not provide an explicit test set.
Hence, for OfficeHome, OfficeHome-RSUT, and VisDA-2017, we use the whole dataset (\ie, trainval set) as the test set following the conventional protocol of UDA and previous work on ADA~\cite{tqs, clue}.

\begin{figure*}[!t]
\begin{center}
\includegraphics[width=0.99 \linewidth]{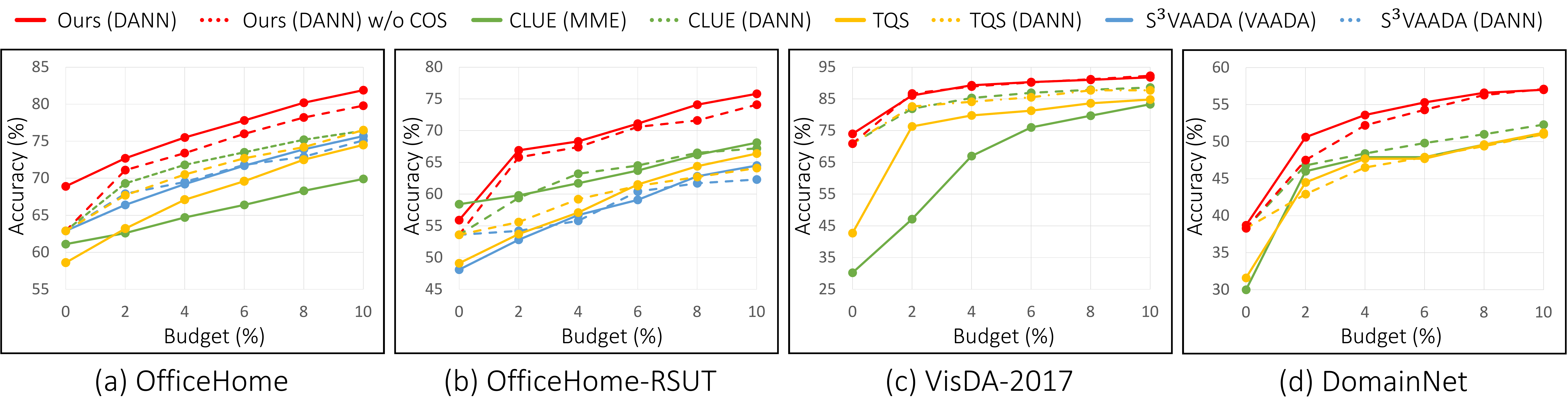}
\end{center}
\vspace{-5mm}
\caption{
Accuracy versus the percent of labeled target instances as budget.
The accuracies are averaged on \textit{all} scenarios of the {OfficeHome}, {OfficeHome-RSUT}, {VisDA-2017}, and {DomainNet}.
The solid lines represent the results of using the specialized adaptation technique of each method, and the dotted lines represent the results of using the same adaptation technique~(\ie, DANN~\cite{dann}).
w$\!$/$\!$o COS: Ours without cosine classifier
}
\label{fig:accuracy}
\vspace{-2mm}
\end{figure*}
\smallskip
\noindent\textbf{Experimental setup.}
We compare LAMDA to the state-of-the-art ADA methods: TQS \cite{tqs}, CLUE \cite{clue},
and $\text{S}^\text{3}$VAADA \cite{s3vaada}.
We note that the existing ADA works have evaluated their methods with different evaluation protocols~(e.g., budget size, sampling interval, and dataset).
For fair comparison, we first benchmark them on four public datasets for domain adaptation through a unified evaluation protocol.
We conduct 5 rounds of data sampling, each of which updates the model from the previous round after newly acquiring labels of 2\%-budget,~\ie, 10\%-budget in total, where we let {\it $n$\%-budget} denote $n$\% of the target train set size.
For both of our method and the previous methods, the model is selected based on the validation accuracy.
For each of the methods, we use the original authors' official implementation.
The detailed descriptions are provided in the supplementary material (\Sec{ada_config}).

\smallskip
\noindent\textbf{Implementation details.}
We use ResNet-50~\cite{resnet} backbone initialized with pretrained weights from ImageNet~\cite{Imagenet} classification for both our and the previous methods.
Our classifier consists of 2 fully connected layers where the embedding dimension $d$ is $512$.
For all experiments, we use an identical set of hyper-parameters.
Our model is trained using SGD optimizer with a learning rate of 0.1, and a weight decay of $5^{-4}$ for 100 epochs.
We set the margin threshold $\Delta$ to 0.8, the temperature $\tau$ in \Eq{predicted_probability} to $0.1$ and the $\gamma$ of RBF kernel in~\Eq{mmd2} to an inverse of the feature dimension, which in our case is $\frac{1}{2048}$.

\subsection{Results} \label{sec:result}

\noindent\textbf{Overall superiority of LAMDA %
with varying budget.}
In~\Fig{accuracy}, we compare the performance of LAMDA and the existing approaches\footnote{
Unfortunately,
$\text{S}^\text{3}$VAADA~\cite{s3vaada} 
for DomainNet and VisDA-2017 
requires
infeasible memory consumption,
in the supplementary material (\Sec{domainnet}), we report its performance 
on a part of scenarios of DomainNet which our resource allows.}
varying budget for each of OfficeHome, OfficeHome-RSUT, VisDa-2017, and DomainNet datasets. 
Note that each method is equipped with its own domain adaptation technique~(\eg, VAADA~\cite{shu2018dirt} for $\text{S}^\text{3}$VAADA, and MME~\cite{mme} for CLUE) and classifier~(\ie, cosine classifier for LAMDA).
We evaluate these methods while varying their adaptation techniques or classifier to examine the contribution of their components thoroughly.
The results show that LAMDA clearly outperforms the previous arts in every setting on all the datasets.
In particular, LAMDA with only $2$\%-budget is often as competitive as or even outperforms the methods with $10$\%-budget.
The performance gap between LAMDA and other methods increases as the budget increases. This suggests that LAMDA utilizes the budget effectively by both ways: label distribution matching and supervised learning.
\begin{table*}[t!]
\centering
\caption{Accuracy (\%)
on {OfficeHome}
using 10\%-budget 
for each source-target pair
of four domains: {\bf A}rt, {\bf C}lipart, {\bf P}roduct, and {\bf R}eal.
w$\!$/$\!$o COS: Ours without cosine classifier
}
\vspace{1mm}
\scalebox{0.7}{
    \begin{tabular}{cl|ccccccccccccc}
            \toprule
            \multirow{2}{*}{\shortstack{DA method}} & \multirow{2}{*}{AL method} & \multicolumn{12}{c}{OfficeHome}  \\
           
            & & A $\veryshortarrow$ C & A $\veryshortarrow$ P & A $\veryshortarrow$ R & C $\veryshortarrow$ A & C $\veryshortarrow$ P & C $\veryshortarrow$ R & P $\veryshortarrow$ A & P $\veryshortarrow$ C & P $\veryshortarrow$ R & R $\veryshortarrow$ A & R $\veryshortarrow$ C & R $\veryshortarrow$ P & Avg  \\
            
            \midrule
           
            - & TQS\cite{tqs}     & 64.3 & 84.8 & 83.5 & 66.1 & 81.0 & 76.7 & 66.5 & 61.4 & 82.0 & 73.7 & 65.9 & 88.5 & 74.5 \\
                        
            MME & CLUE\cite{clue}    & 62.1 & 80.6 & 73.9 & 55.2 & 76.4 & 75.4 & 53.9 & 62.1 & 80.7 & 67.5 & 63.0 & 88.1 & 69.9 \\
            
            VAADA & $\text{S}^\text{3}$VAADA\cite{s3vaada} & 67.8 & 83.9 & 82.9 & 67.0 & 81.5 & 79.5 & 65.8 & 65.9 & 82.4 & 74.8 & 68.6 & 87.8 & 75.7 \\       
            
            \midrule
            
            \multirow{5}{*}{DANN}
            & TQS\cite{tqs}                      & 68.7 & 80.1 & 83.1 & 64.0 & 83.1 & 76.9 & 67.7 & 71.0 & 84.4 & 76.4 & 72.7 & 90.0 & 76.5 \\
                        
            & CLUE\cite{clue}                     & 70.3 & 81.9 & 80.4 & 65.6 & 83.8 & 75.8 & 64.7 & 73.9 & 82.7 & 76.1 & 74.3 & 87.0 & 76.4 \\
            
            & $\text{S}^\text{3}$VAADA\cite{s3vaada} & 65.5 & 79.6 & 80.0 & 65.4 & 82.2 & 75.5 & 68.4 & 68.1 & 84.0 & 73.5 & 70.7 & 88.6 & 75.1 \\
            & Ours w$\!$/$\!$o COS      & \underline{73.0} & \underline{87.6} & \underline{84.2} & \underline{69.5} & \underline{85.9} & \underline{81.0} & \underline{71.9} & \underline{74.6} & \underline{85.3} & \underline{77.3} & \underline{75.9} & \underline{91.6} & \underline{79.8} \\
            & Ours       & \textbf{74.8} & \textbf{88.5} & \textbf{86.9} & \textbf{73.8} & \textbf{88.2} & \textbf{83.3} & \textbf{74.6} & \textbf{75.5} & \textbf{86.9} & \textbf{80.8} & \textbf{77.8} & \textbf{91.7} & \textbf{81.9} \\
            \bottomrule
    \end{tabular}
    }
    \label{tab:office_home}
\end{table*}
\begin{table}[t!]
    \caption{(a) Accuracy (\%) on {OfficeHome-RSUT} using 10\%-budget for each source-target pair of three domains: {\bf C}lipart, {\bf P}roduct, and {\bf R}eal.
    (b) Accuracy (\%) on {VisDA-2017} and {DomainNet} using 10\%-budget where VisDA-2017 consists of two domains: {\bf R}eal and {\bf S}ynthetic, and DomainNet consists of five domains: {\bf R}eal, {\bf C}lipart, {\bf S}ketch, {\bf P}ainting, and {\bf Q}uickdraw.
    w$\!$/$\!$o COS: Ours without cosine classifier}
    \vspace{1mm}
    \begin{subtable}[h]{0.48\textwidth}
        \centering
        \renewcommand{\arraystretch}{1.15}
        \scalebox{0.51}{
        \begin{tabular}{cl|ccccccc}
                    \toprule
                    \multirow{2}{*}{\shortstack{DA method}} & \multirow{2}{*}{AL method} & \multicolumn{7}{c}{OfficeHome-RSUT}  \\
                   
                    & & C $\veryshortarrow$ P & C $\veryshortarrow$ R & P $\veryshortarrow$ C & P $\veryshortarrow$ R & R $\veryshortarrow$ C & R $\veryshortarrow$ P &  Avg  \\
                    
                    \midrule
                   
                    - & TQS\cite{tqs} & 69.4 & 65.7 & 53.0 & 76.3 & 53.1 & 81.1 & 66.4 \\
                                                                    
                    MME & CLUE\cite{clue} & 69.7 & 65.9 & 57.1 & 73.4 & 59.5 & 82.7 & 68.1 \\
                    
                    VAADA & $\text{S}^\text{3}$VAADA\cite{s3vaada} & 73.0 & 63.0 & 50.7 & 69.6 & 52.6 & 78.3 & 64.5 \\            
                    
                    \midrule
                    
                    \multirow{5}{*}{DANN}
                    & TQS\cite{tqs}             & 67.6 & 61.4 & 54.8 & 74.7 & 53.6 & 77.6 & 64.9 \\
                                
                    & CLUE\cite{clue}           & 71.5 & 64.3 & 56.3 & 76.5 & 54.6 & 79.9 & 67.2 \\
                    
                    & $\text{S}^\text{3}$VAADA\cite{s3vaada}  & 66.9 & 61.4 & 53.0 & 75.4 & 52.4 & 76.4 & 64.2 \\
                    & Ours w$\!$/$\!$o COS             & \underline{78.1} & \underline{72.1} & \underline{61.5} & \textbf{82.3} & \underline{64.2} & \underline{86.5} & \underline{74.1} \\
                    & Ours                     & \textbf{81.2} & \textbf{75.7} & \textbf{64.1} & \underline{81.6} & \textbf{65.1} & \textbf{87.2} & \textbf{75.8} \\
              
                    \bottomrule
        \end{tabular}}
        \caption{}
        \label{tab:office_home_rsut}
    \end{subtable}
    \hfill
    \begin{subtable}[h]{0.5\textwidth}
        \centering
        \large
        \renewcommand{\arraystretch}{1.251}
        \scalebox{0.5}{
        \begin{tabular}{cl|cccccc}
                    \toprule
                    \multirow{2}{*}{\shortstack{DA method}} & \multirow{2}{*}{AL method} & VisDa-2017 & \multicolumn{5}{c}{DomainNet} \\
                   
                    & & S $\veryshortarrow$ R & R $\veryshortarrow$ C & C $\veryshortarrow$ S & S $\veryshortarrow$ P & C $\veryshortarrow$ Q & Avg \\
                    
                    \midrule
                   
                    - & TQS\cite{tqs}    & 84.8 & 54.2 & 51.7 & 51.4 & 47.4 & 51.2 \\
                                                                    
                    MME & CLUE\cite{clue}    & 83.3 & 60.7 & 50.4 & 53.5 & 39.4 & 51.0 \\
                                
                    \midrule
                    
                    \multirow{4}{*}{DANN}
                    & TQS\cite{tqs}    & 87.7 & 59.3 & 50.9 & 52.4 & 41.5 & 51.0 \\
                    
                    & CLUE\cite{clue}    & 88.6 & 60.9 & 52.2 & 52.4 & 43.7 & 52.3 \\
                    
                    & Ours w$\!$/$\!$o COS   & \textbf{92.3} & \underline{64.6} & \textbf{56.4} & \textbf{58.7} & \textbf{48.5} & \textbf{57.1} \\ 

                    & Ours    & \underline{91.8} & \textbf{65.3} & \underline{56.1} & \underline{58.1} & \underline{48.3} & \underline{57.0} \\ 
                    \bottomrule
        \end{tabular}
        }
        \caption{}
        \label{tab:domainnet_visda}
    \end{subtable}
    \vspace{-3mm}
     \label{tab:rsut_domiannet_visda}
\end{table}

\smallskip
\noindent\textbf{Advantages of LAMDA across diverse source-target domain pairs.}
In Table~\ref{tab:office_home}-\ref{tab:rsut_domiannet_visda}, we compare LAMDA and the existing ADA methods in every domain adaptation scenario of the four datasets given 10\%-budget, where LAMDA always outperforms the others.
Regarding that OfficeHome-RSUT has a significant class distribution shift compared to OfficeHome, the advantage of LAMDA equipped with the label distribution matching becomes clearer in OfficeHome-RSUT (Table~\ref{tab:office_home_rsut}) than OfficeHome (Table~\ref{tab:office_home}).
Table~\ref{tab:domainnet_visda} demonstrates the scalability of LAMDA,
where it clearly outperforms the previous work by about 4\% or more in all scenarios of the large-scale datasets, VisDA 2017 and DomainNet.
In the supplementary material (\Sec{ssda}),
we also show that LAMDA surpasses state-of-the-art SSDA methods~\cite{cdac,ecacl}.

\subsection{Analysis} \label{sec:analysis}

\begin{table}
\centering
\caption{Accuracy (\%) averaged over \textit{all} scenarios when using 10\%-budget, where we conduct an ablation study
from ablation baseline at the last row to LAMDA at the first row by sequentially adding three components:
(i) Prototype: sampling described in \Sec{prototype} (o/w, sampling uniformly at random);
(ii) Matching: label distribution matching in \Sec{calibrated_dann} (o/w, replacing $p_i$ in \Eq{src_sampling} with uniform distribution); and
(iii) Cosine: cosine classifier described in \Sec{training} (o/w, linear classifier).
($\cdot$): accuracy gain by adding each component.
}
\vspace{1mm}
\renewcommand{\arraystretch}{1.1}
\scalebox{0.8}{
\begin{tabular}{ccc|cc}
            \toprule
            Prototype & Matching & Cosine & OfficeHome & OfficeHome-RSUT \\
                       
            \midrule
            
            \cmark & \cmark & \cmark & 81.9 (+2.1) & 75.8 (+1.7) \\
                                                
            \cmark & \cmark & \xmark & 79.8 (+2.7) & 74.1 (+6.8) \\
            
            \cmark & \xmark & \xmark & 77.1 (+3.8) & 67.3 (+3.7) \\
            
            \xmark & \xmark & \xmark & 73.3 & 63.6 \\
            \bottomrule
\end{tabular}}
\label{tab:analysis}
\end{table}
\noindent\textbf{Contribution of each component of LAMDA.}
\Tbl{analysis} quantifies the contribution of each components of LAMDA:
(i) prototype set sampling in \Sec{prototype};
(ii) label distribution matching in \Sec{calibrated_dann};
and (iii) cosine classifier in \Sec{training}.
Every component in LAMDA improves the performance in both OfficeHome and OfficeHome-RSUT.
The performance gap between the last (random sampling with DANN~\cite{dann}) and the second last rows verifies that our prototype sampling method boost the effect of supervised learning.
Comparing the second and the third rows, one can see the remarkable performance gain by our label distribution matching strategy, in particular on OfficeHome-RSUT with significant label distribution shift.
Finally, the use of cosine classifier further improves performance by 1.9\% in average.
The results for every individual adaptation scenarios are reported in the supplementary material (\Sec{component}).

\begin{figure}[!t]
\begin{center}
\includegraphics[width=1.0 \linewidth]{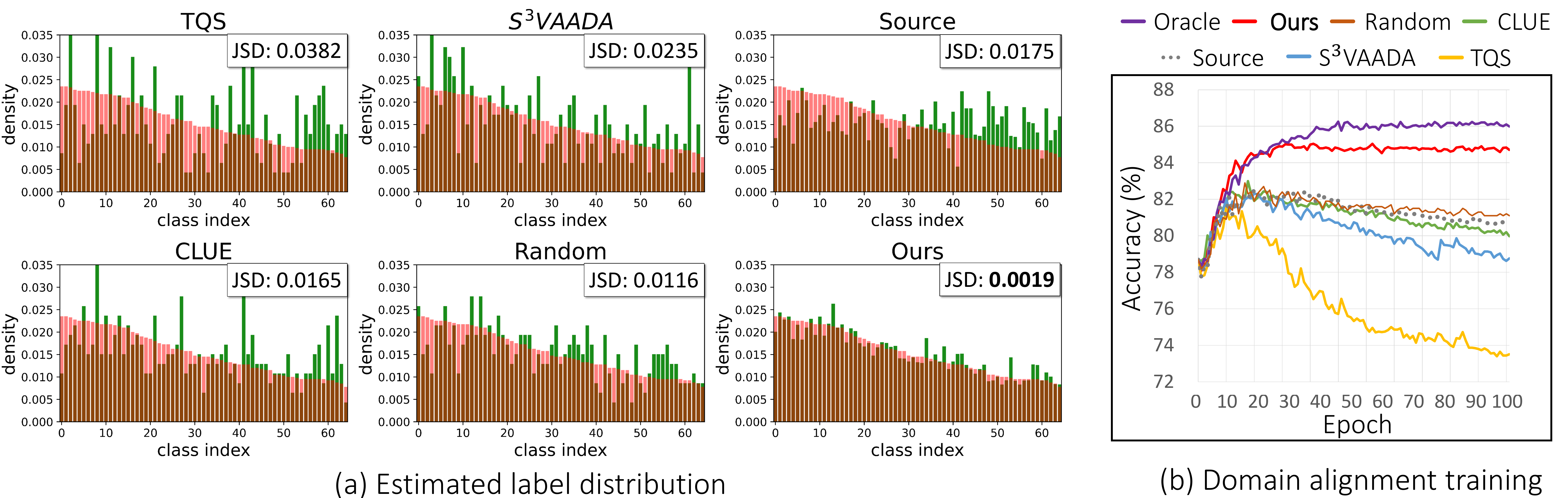}
\end{center}
\vspace{-4.4mm}
\caption{
Effect of active sampling strategy on label distribution matching (\Sec{calibrated_dann}) in OfficeHome Real to Product scenario.
(a) The true (red) and the estimated (green) label distribution of target domain, where each sampling methods estimates the distribution using 10\%-budget.
The methods are sorted by the estimation quality.
JSD: Jensen-Shannon Divergence between the estimated and the true label distribution (lower is better).
Source: Label distribution of source data.
(b) Training curve of domain alignment learning (\Eq{dann}) combined with label distribution matching using the estimations in (a).
Source: naive domain alignment.
Oracle: using true target label distribution.
}
\label{fig:jsd_acc}
\vspace{-2mm}
\end{figure}

\smallskip
\noindent\textbf{Quality of estimated label distribution.}
As described in \Sec{calibrated_dann}, estimating target label distribution plays a prominent role in LAMDA.
In~\Fig{jsd_acc}a, we visualize label distributions of sampled data of LAMDA and those of the previous work, and compute Jensen-Shannon divergence (JSD) between the estimated distributions and the true one.
The results demonstrate that LAMDA enables to estimate target label distribution most accurately compared to the previous work and the random sampling, which is a naive but intuitive sampling strategy for the estimation.
Note that the previous work is even worse than the random sampling in terms of the estimation accuracy,
which empirically reconfirm that the sampling strategies of the previous work are not aware of the target data distribution.
When solely utilizes all of the pseudo-labels from source pretrained model for the estimation, it gives JSD of 0.025 which is worse than `Source' baseline.
This is mainly due to the bias of the pseudo-labeled data; they are highly confident samples.
Our sampling method avoids this bias by combining labeled and pseudo-labeled data.

\noindent\textbf{Benefit of label distribution matching.}
In~\Fig{jsd_acc}b, we plot training curves of domain alignment combined with label distribution matching,
where each methods utilizes identical source classification loss and domain alignment loss as in~\Eq{dann}, but with different label distribution estimated from each sampling methods in~\Fig{jsd_acc}a.
Training without label distribution matching (\eg, \emph{Source}) or matching with inaccurate target label distribution degrade accuracy, while ours does not, thanks to the accurate estimation of target label distribution.
It is worth noting that our model using 10\%-budget shows comparable accuracy with \emph{Oracle}, which has access to the true target label distribution.

\begin{figure*}[!t]
\begin{center}
\includegraphics[width=0.98\linewidth]{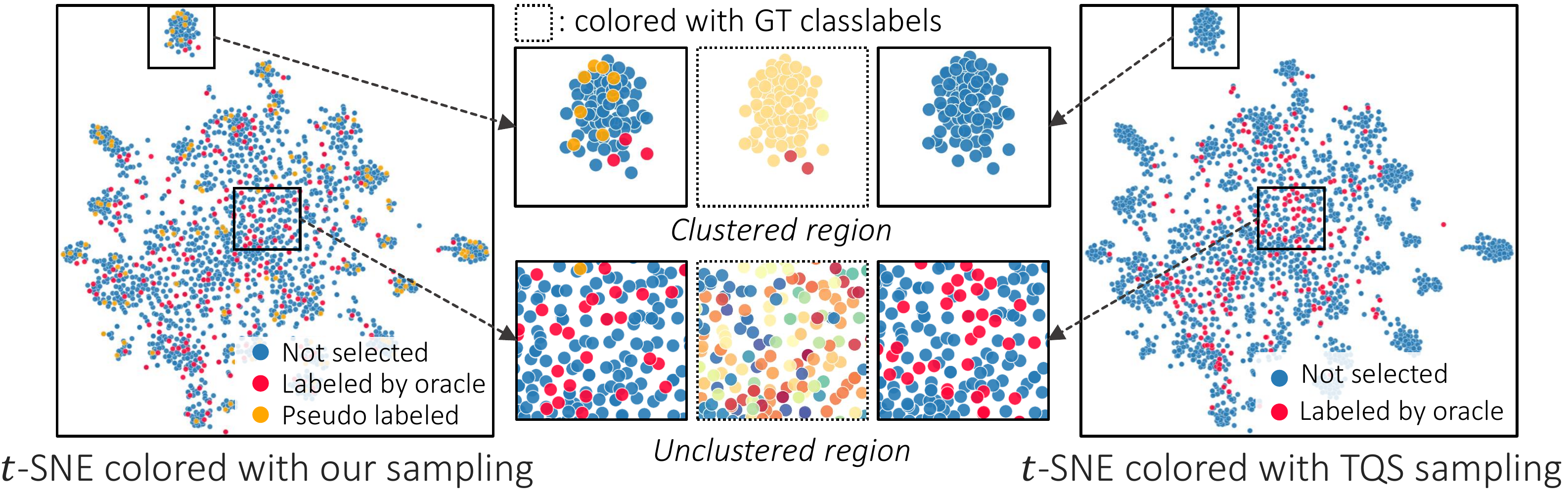}
\end{center}
\vspace{-2mm}
\caption{
$t$-SNE~\cite{tsne} visualization of target feature vectors from source pre-trained model on OfficeHome Real to Art scenario.}
\label{fig:tsne}
\vspace{-2mm}
\end{figure*}

\noindent\textbf{Visualization of sampled data by $t$-SNE.}
\Fig{tsne} visualizes distributions of target features and those selected by LAMDA and TQS, to show the difference of their sampling strategies.
Since TQS prefers to select uncertain data, mostly located in unclustered regions, its samples do not reflect the target data distribution, \eg, the certain instances in the clustered region are undersampled.
In contrast, LAMDA considers certain samples ignored in TQS and assigns them pseudo-labels for label distribution prediction, while it requests an oracle to annotate uncertain data within the budget.
Such a sampling strategy allows us to mainly invest a budget on uncertain data while utilizing density-aware samples to estimate the target label distribution.
These observations align with our design rationale, depicted in~\Fig{sampling}.
We also visualize the selected target feature vectors of CLUE and $\text{S}^\text{3}$VAADA in the supplementary material (\Sec{supp_tsne}).

\begin{wrapfigure}{r}{0.48\textwidth}
\vspace{-3.5mm}
\begin{center}
\includegraphics[width = 0.46\textwidth]{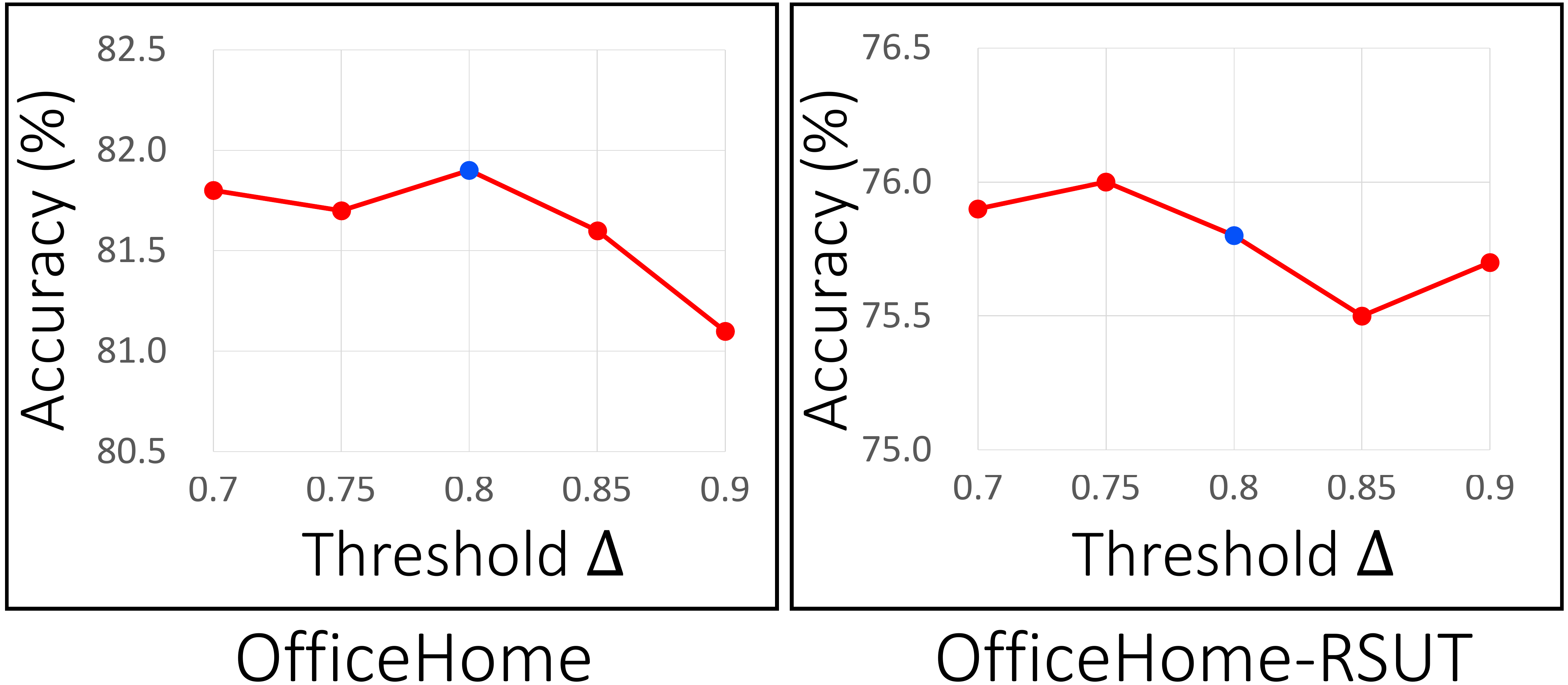}
\end{center}
\vspace{-6mm}
\caption{
Accuracy of LAMDA versus hyper-parameter $\Delta$.
The blue dot indicates the value used in the main paper.}
\label{fig:delta}
\vspace{-7mm}
\end{wrapfigure}
\smallskip
\noindent\textbf{Hyper-parameter analysis.}
In~\Fig{delta}, we evaluate the sensitivity of LAMDA to the choice of the threshold $\Delta$ in~\Alg{sampling}.
LAMDA is surprisingly robust to the change of $\Delta$, where the change of accuracy is less than 1\% for both OfficeHome and OfficeHome-RSUT when the $\Delta$ is between 0.7 and 0.9.
We note that while the optimal value of $\Delta$ varies among the datasets, we use the same value for all of our experiments.
When we do not utilize the pseudo label in LAMDA (\ie, $\Delta = 1$), the accuracy drops 4\% and 1.1\% in OfficeHome and OfficeHome-RSUT, respectively.
This shows the effectiveness of our prototype sampling strategy.

\begin{wrapfigure}{r}{0.5\textwidth}
\vspace{-12mm}
\begin{center}
\includegraphics[width = 0.42\textwidth]{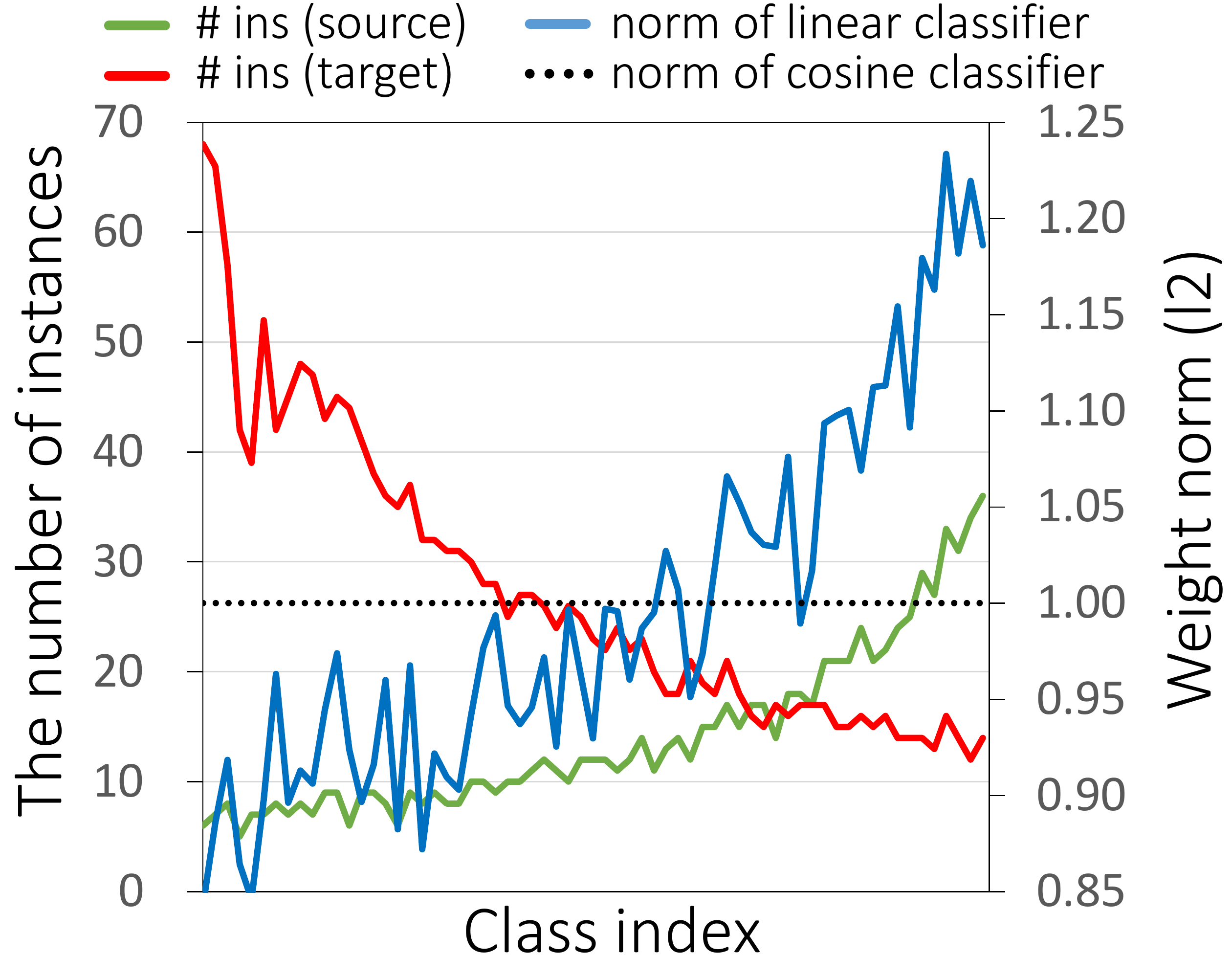}
\end{center}
\vspace{-7mm}
\caption{
The frequencies (\texttt{\#} ins) of each class in source and target domain (OfficeHome-RSUT Clipart to Product scenario) and the l2-norm of the corresponding classifier weight trained with the source data.} 
\label{fig:weight_norm}
\vspace{-7mm}
\end{wrapfigure}
\smallskip
\noindent\textbf{Analysis of cosine classifier.}
To inspect the cosine classifier, we compare in~\Fig{weight_norm} the frequencies and the weight norm of the linear classifier for each class.
The norm of the linear classifier is positively correlated to the frequencies of each class within the source domain (blue and green lines).
Since a large norm of classifier weights has been known to result in predictions biased to major classes~\cite{Kang2020Decoupling},
the mismatch between the classifier norm and the target domain class frequencies
(red and blue lines) is undesirable.
The cosine classifier alleviates this issue by normalizing its weight scale.
In the supplementary material (\Sec{supp_cosine}), we evaluate existing ADA methods combined with cosine classifier.
\section{Conclusion}
We proposed LAMDA, a new method to address the issue of label distribution shift in ADA.
It selects target data best preserving the target data distribution as well as being representative, diverse, and uncertain.
During training, LAMDA estimates the label distribution of the target domain, and builds each source data mini-batch in a way that the label frequencies of the batch follow the estimated target label distribution.
On the four different domain adaptation datasets, the proposed method substantially outperforms all the existing ADA models.

\vspace{2mm}
{\small
\noindent \textbf{Acknowledgement.} 
We sincerely appreciate Moonjeong Park for fruitful discussions.
This work was supported by 
the NRF grant and  %
the IITP grant %
funded by Ministry of Science and ICT, Korea
(NRF-2018R1A5-A1060031, %
 NRF-2021R1A2C3012728,  %
 IITP-2019-0-01906,     %
 IITP-2020-0-00842,     %
 IITP-2021-0-02068,     %
 IITP-2022-0-00290).    %
}
\pagebreak

\clearpage
{\small
\bibliographystyle{splncs04}
\bibliography{cvlab_kwak}

\begin{thebibliography}{10}
\providecommand{\url}[1]{\texttt{#1}}
\providecommand{\urlprefix}{URL }
\providecommand{\doi}[1]{https://doi.org/#1}

\bibitem{ajakan2014domain}
Ajakan, H., Germain, P., Larochelle, H., Laviolette, F., Marchand, M.:
  Domain-adversarial neural networks. arXiv preprint arXiv:1412.4446  (2014)

\bibitem{asghar2016deep}
Asghar, N., Poupart, P., Jiang, X., Li, H.: Deep active learning for dialogue
  generation. In: Proceedings of the 6th Joint Conference on Lexical and
  Computational Semantics (*{SEM} 2017) (2017)

\bibitem{ash2019deep}
Ash, J.T., Zhang, C., Krishnamurthy, A., Langford, J., Agarwal, A.: Deep batch
  active learning by diverse, uncertain gradient lower bounds. In: Proc.
  International Conference on Learning Representations (ICLR) (2020)

\bibitem{ben2010theory}
Ben-David, S., Blitzer, J., Crammer, K., Kulesza, A., Pereira, F., Vaughan,
  J.W.: A theory of learning from different domains. Machine learning
  \textbf{79}(1),  151--175 (2010)

\bibitem{chen2018a}
Chen, W.Y., Liu, Y.C., Kira, Z., Wang, Y.C.F., Huang, J.B.: A closer look at
  few-shot classification. In: Proc. International Conference on Learning
  Representations (ICLR) (2019)

\bibitem{tachet2020domain}
Tachet~des Combes, R., Zhao, H., Wang, Y.X., Gordon, G.J.: Domain adaptation
  with conditional distribution matching and generalized label shift. In: Proc.
  Neural Information Processing Systems (NeurIPS) (2020)

\bibitem{courty2017joint}
Courty, N., Flamary, R., Habrard, A., Rakotomamonjy, A.: Joint distribution
  optimal transportation for domain adaptation. In: Proc. Neural Information
  Processing Systems (NeurIPS) (2017)

\bibitem{courty2016optimal}
Courty, N., Flamary, R., Tuia, D., Rakotomamonjy, A.: Optimal transport for
  domain adaptation. IEEE Transactions on Pattern Analysis and Machine
  Intelligence (TPAMI)  \textbf{39}(9),  1853--1865 (2016)

\bibitem{gvb}
Cui, S., Wang, S., Zhuo, J., Su, C., Huang, Q., Tian, Q.: Gradually vanishing
  bridge for adversarial domain adaptation. In: Proc. Neural Information
  Processing Systems (NeurIPS) (2020)

\bibitem{dai2020curriculum}
Dai, D., Sakaridis, C., Hecker, S., Van~Gool, L.: Curriculum model adaptation
  with synthetic and real data for semantic foggy scene understanding.
  International Journal of Computer Vision (IJCV)  (2020)

\bibitem{damodaran2018deepjdot}
Damodaran, B.B., Kellenberger, B., Flamary, R., Tuia, D., Courty, N.: Deepjdot:
  Deep joint distribution optimal transport for unsupervised domain adaptation.
  In: Proceedings of the European Conference on Computer Vision (ECCV) (2018)

\bibitem{deheeger2021discrepancy}
Deheeger, F., MOUGEOT, M., Vayatis, N., et~al.: Discrepancy-based active
  learning for domain adaptation. In: Proc. International Conference on
  Learning Representations (ICLR) (2021)

\bibitem{Imagenet}
Deng, J., Dong, W., Socher, R., Li, L.J., Li, K., Fei-Fei, L.: {ImageNet:} a
  large-scale hierarchical image database. In: Proc. IEEE/CVF Conference on
  Computer Vision and Pattern Recognition (CVPR) (2009)

\bibitem{tqs}
Fu, B., Cao, Z., Wang, J., Long, M.: Transferable query selection for active
  domain adaptation. In: Proc. IEEE/CVF Conference on Computer Vision and
  Pattern Recognition (CVPR) (2021)

\bibitem{dann}
Ganin, Y., Ustinova, E., Ajakan, H., Germain, P., Larochelle, H., Laviolette,
  F., Marchand, M., Lempitsky, V.: Domain-adversarial training of neural
  networks. Journal of Machine Learning Research (JMLR)  \textbf{17}(1),
  2096–2030 (2016)

\bibitem{gidaris2018dynamic}
Gidaris, S., Komodakis, N.: Dynamic few-shot visual learning without
  forgetting. In: Proc. IEEE/CVF Conference on Computer Vision and Pattern
  Recognition (CVPR) (2018)

\bibitem{mmd}
Gretton, A., Borgwardt, K.M., Rasch, M.J., Sch{\"o}lkopf, B., Smola, A.: A
  kernel two-sample test. Journal of Machine Learning Research (JMLR)
  \textbf{13}(1),  723--773 (2012)

\bibitem{resnet}
He, K., Zhang, X., Ren, S., Sun, J.: Deep residual learning for image
  recognition. In: Proc. IEEE/CVF Conference on Computer Vision and Pattern
  Recognition (CVPR) (2016)

\bibitem{he2019towards}
He, T., Jin, X., Ding, G., Yi, L., Yan, C.: Towards better uncertainty
  sampling: Active learning with multiple views for deep convolutional neural
  network. In: 2019 IEEE International Conference on Multimedia and Expo (ICME)
  (2019)

\bibitem{huang2005automated}
Huang, J.Z., Ng, M.K., Rong, H., Li, Z.: Automated variable weighting in
  k-means type clustering. IEEE Transactions on Pattern Analysis and Machine
  Intelligence (TPAMI)  \textbf{27}(5),  657--668 (2005)

\bibitem{Batchnorm}
Ioffe, S., Szegedy, C.: Batch normalization: Accelerating deep network training
  by reducing internal covariate shift. In: Proc. International Conference on
  Machine Learning (ICML) (2015)

\bibitem{dalib}
Jiang, J., Chen, B., Fu, B., Long, M.: Transfer-learning-library.
  \url{https://github.com/thuml/Transfer-Learning-Library} (2020)

\bibitem{jin2020minimum}
Jin, Y., Wang, X., Long, M., Wang, J.: Minimum class confusion for versatile
  domain adaptation. In: European Conference on Computer Vision. pp. 464--480.
  Springer (2020)

\bibitem{Kang2020Decoupling}
Kang, B., Xie, S., Rohrbach, M., Yan, Z., Gordo, A., Feng, J., Kalantidis, Y.:
  Decoupling representation and classifier for long-tailed recognition. In:
  Proc. International Conference on Learning Representations (ICLR) (2020)

\bibitem{mmdcritic}
Kim, B., Khanna, R., Koyejo, O.O.: Examples are not enough, learn to criticize!
  criticism for interpretability. In: Proc. Neural Information Processing
  Systems (NeurIPS) (2016)

\bibitem{lecun2006tutorial}
LeCun, Y., Chopra, S., Hadsell, R., Ranzato, M., Huang, F.: A tutorial on
  energy-based learning. Predicting structured data  \textbf{1}(0) (2006)

\bibitem{lee2019sliced}
Lee, C.Y., Batra, T., Baig, M.H., Ulbricht, D.: Sliced wasserstein discrepancy
  for unsupervised domain adaptation. In: Proc. IEEE/CVF Conference on Computer
  Vision and Pattern Recognition (CVPR) (2019)

\bibitem{lee2022fifo}
Lee, S., Son, T., Kwak, S.: Fifo: Learning fog-invariant features for foggy
  scene segmentation. In: Proc. IEEE/CVF Conference on Computer Vision and
  Pattern Recognition (CVPR) (2022)

\bibitem{cdac}
Li, J., Li, G., Shi, Y., Yu, Y.: Cross-domain adaptive clustering for
  semi-supervised domain adaptation. In: Proc. IEEE/CVF Conference on Computer
  Vision and Pattern Recognition (CVPR) (2021)

\bibitem{ecacl}
Li, K., Liu, C., Zhao, H., Zhang, Y., Fu, Y.: Ecacl: A holistic framework for
  semi-supervised domain adaptation. In: Proc. IEEE/CVF International
  Conference on Computer Vision (ICCV) (2021)

\bibitem{mkmmd}
Long, M., Cao, Y., Wang, J., Jordan, M.: Learning transferable features with
  deep adaptation networks. In: Proc. International Conference on Machine
  Learning (ICML). PMLR (2015)

\bibitem{cdan}
Long, M., Cao, Z., Wang, J., Jordan, M.I.: Conditional adversarial domain
  adaptation. Proc. Neural Information Processing Systems (NeurIPS)  (2018)

\bibitem{long2017deep}
Long, M., Zhu, H., Wang, J., Jordan, M.I.: Deep transfer learning with joint
  adaptation networks. In: Proc. International Conference on Machine Learning
  (ICML). PMLR (2017)

\bibitem{tsne}
Van~der Maaten, L., Hinton, G.: Visualizing data using t-sne. Journal of
  Machine Learning Research (JMLR)  \textbf{9}(11) (2008)

\bibitem{nemhauser1978analysis}
Nemhauser, G.L., Wolsey, L.A., Fisher, M.L.: An analysis of approximations for
  maximizing submodular set functions—i. Mathematical programming
  \textbf{14}(1),  265--294 (1978)

\bibitem{ostapuk2019activelink}
Ostapuk, N., Yang, J., Cudr{\'e}-Mauroux, P.: Activelink: deep active learning
  for link prediction in knowledge graphs. In: The World Wide Web Conference
  (WWW) (2019)

\bibitem{domainnet}
Peng, X., Bai, Q., Xia, X., Huang, Z., Saenko, K., Wang, B.: Moment matching
  for multi-source domain adaptation. In: Proc. IEEE/CVF International
  Conference on Computer Vision (ICCV) (2019)

\bibitem{peng2018visda}
Peng, X., Usman, B., Kaushik, N., Wang, D., Hoffman, J., Saenko, K.: Visda: A
  synthetic-to-real benchmark for visual domain adaptation. In: Proceedings of
  the IEEE Conference on Computer Vision and Pattern Recognition Workshops. pp.
  2021--2026 (2018)

\bibitem{clue}
Prabhu, V., Chandrasekaran, A., Saenko, K., Hoffman, J.: Active domain
  adaptation via clustering uncertainty-weighted embeddings. In: Proc. IEEE/CVF
  International Conference on Computer Vision (ICCV) (2021)

\bibitem{purushotham2016variational}
Purushotham, S., Carvalho, W., Nilanon, T., Liu, Y.: Variational recurrent
  adversarial deep domain adaptation. In: Proc. International Conference on
  Learning Representations (ICLR) (2016)

\bibitem{qi2018low}
Qi, H., Brown, M., Lowe, D.G.: Low-shot learning with imprinted weights. In:
  Proc. IEEE/CVF Conference on Computer Vision and Pattern Recognition (CVPR)
  (2018)

\bibitem{rai2010domain}
Rai, P., Saha, A., Daum{\'e}~III, H., Venkatasubramanian, S.: Domain adaptation
  meets active learning. In: Proceedings of the NAACL HLT 2010 Workshop on
  Active Learning for Natural Language Processing (2010)

\bibitem{s3vaada}
Rangwani, H., Jain, A., Aithal, S.K., Babu, R.V.: S3vaada: Submodular subset
  selection for virtual adversarial active domain adaptation. In: Proc.
  IEEE/CVF International Conference on Computer Vision (ICCV) (2021)

\bibitem{rdusseeun1987clustering}
Rdusseeun, L., Kaufman, P.: Clustering by means of medoids. In: Proceedings of
  the statistical data analysis based on the L1 norm conference, neuchatel,
  switzerland. vol.~31 (1987)

\bibitem{roth2006margin}
Roth, D., Small, K.: Margin-based active learning for structured output spaces.
  In: European Conference on Machine Learning. pp. 413--424. Springer (2006)

\bibitem{mme}
Saito, K., Kim, D., Sclaroff, S., Darrell, T., Saenko, K.: Semi-supervised
  domain adaptation via minimax entropy. In: Proc. IEEE/CVF International
  Conference on Computer Vision (ICCV) (2019)

\bibitem{sakaridis2019guided}
Sakaridis, C., Dai, D., Gool, L.V.: Guided curriculum model adaptation and
  uncertainty-aware evaluation for semantic nighttime image segmentation. In:
  Proc. IEEE/CVF International Conference on Computer Vision (ICCV) (2019)

\bibitem{Sakaridis_2018_ECCV}
Sakaridis, C., Dai, D., Hecker, S., Van~Gool, L.: Model adaptation with
  synthetic and real data for semantic dense foggy scene understanding. In:
  Proc. European Conference on Computer Vision (ECCV) (2018)

\bibitem{Sakaridis_2018_IJCV}
Sakaridis, C., Dai, D., Van~Gool, L.: Semantic foggy scene understanding with
  synthetic data. International Journal of Computer Vision (IJCV)  (2018)

\bibitem{sener2017active}
Sener, O., Savarese, S.: Active learning for convolutional neural networks: A
  core-set approach. In: Proc. International Conference on Learning
  Representations (ICLR) (2018)

\bibitem{settles2009active}
Settles, B.: Active learning literature survey. Computer Sciences Technical
  Report~1648, University of Wisconsin--Madison (2009)

\bibitem{seung1992query}
Seung, H.S., Opper, M., Sompolinsky, H.: Query by committee. In: Proceedings of
  the fifth annual workshop on Computational learning theory. pp. 287--294
  (1992)

\bibitem{shu2018dirt}
Shu, R., Bui, H.H., Narui, H., Ermon, S.: A dirt-t approach to unsupervised
  domain adaptation. In: Proc. International Conference on Learning
  Representations (ICLR) (2018)

\bibitem{sinha2019variational}
Sinha, S., Ebrahimi, S., Darrell, T.: Variational adversarial active learning.
  In: Proc. IEEE/CVF International Conference on Computer Vision (ICCV) (2019)

\bibitem{su2020active}
Su, J.C., Tsai, Y.H., Sohn, K., Liu, B., Maji, S., Chandraker, M.: Active
  adversarial domain adaptation. In: Proc. IEEE/CVF Winter Conference on
  Applications of Computer Vision (WACV) (2020)

\bibitem{tan2020class}
Tan, S., Peng, X., Saenko, K.: Class-imbalanced domain adaptation: an empirical
  odyssey. In: Proc. European Conference on Computer Vision (ECCV) (2020)

\bibitem{tzeng2015simultaneous}
Tzeng, E., Hoffman, J., Darrell, T., Saenko, K.: Simultaneous deep transfer
  across domains and tasks. In: Proc. IEEE/CVF International Conference on
  Computer Vision (ICCV) (2015)

\bibitem{officehome}
Venkateswara, H., Eusebio, J., Chakraborty, S., Panchanathan, S.: Deep hashing
  network for unsupervised domain adaptation. In: Proc. IEEE/CVF Conference on
  Computer Vision and Pattern Recognition (CVPR) (2017)

\bibitem{wang2019incorporating}
Wang, Z., Du, B., Tu, W., Zhang, L., Tao, D.: Incorporating distribution
  matching into uncertainty for multiple kernel active learning. IEEE
  Transactions on Knowledge and Data Engineering  \textbf{33}(1),  128--142
  (2019)

\bibitem{wang2015querying}
Wang, Z., Ye, J.: Querying discriminative and representative samples for batch
  mode active learning. ACM Transactions on Knowledge Discovery from Data
  (TKDD)  \textbf{9}(3),  1--23 (2015)

\bibitem{xie2022active}
Xie, B., Yuan, L., Li, S., Liu, C.H., Cheng, X., Wang, G.: Active learning for
  domain adaptation: An energy-based approach. In: Proc. AAAI Conference on
  Artificial Intelligence (AAAI) (2022)

\bibitem{xie2022learning}
Xie, M., Li, Y., Wang, Y., Luo, Z., Gan, Z., Sun, Z., Chi, M., Wang, C., Wang,
  P.: Learning distinctive margin toward active domain adaptation. In: Proc.
  IEEE/CVF Conference on Computer Vision and Pattern Recognition (CVPR) (2022)

\bibitem{xu2019larger}
Xu, R., Li, G., Yang, J., Lin, L.: Larger norm more transferable: An adaptive
  feature norm approach for unsupervised domain adaptation. In: Proc. IEEE/CVF
  International Conference on Computer Vision (ICCV) (2019)

\bibitem{decota}
Yang, L., Wang, Y., Gao, M., Shrivastava, A., Weinberger, K.Q., Chao, W.L.,
  Lim, S.N.: Deep co-training with task decomposition for semi-supervised
  domain adaptation. In: Proc. IEEE/CVF International Conference on Computer
  Vision (ICCV) (2021)

\bibitem{yoon2022semi}
Yoon, J., Kang, D., Cho, M.: Semi-supervised domain adaptation via
  sample-to-sample self-distillation. In: Proceedings of the IEEE/CVF Winter
  Conference on Applications of Computer Vision. pp. 1978--1987 (2022)

\bibitem{zhao19}
Zhao, H., Des~Combes, R.T., Zhang, K., Gordon, G.: On learning invariant
  representations for domain adaptation. In: Proc. International Conference on
  Machine Learning (ICML). PMLR (2019)

\end{thebibliography}
}

\clearpage

\title{Combating Label Distribution Shift for\\Active Domain Adaptation\\ %
{\it  ---Supplementary Material--- }}
\author{Sehyun Hwang\inst{1} \qquad
Sohyun Lee\inst{2} \qquad
Sungyeon Kim\inst{1} \qquad
\\
Jungseul Ok\inst{1,2}\thanks{Co-corresponding authors} \qquad
Suha~Kwak\inst{1,2}$^\star$
}
\authorrunning{Sehyun Hwang, Sohyun Lee, Sungyeon Kim, Jungseul Ok, and Suha Kwak}
\institute{
$^{1}$Department of Computer Science and Engineering, POSTECH, Korea\\
$^{2}$Graduate School of Artificial Intelligence, POSTECH, Korea\\
}
\maketitle

\appendix
\renewcommand{\theequation}{a\arabic{equation}}
\renewcommand{\thetable}{a\arabic{table}}
\renewcommand{\thefigure}{a\arabic{figure}}

\noindent This material provides additional analysis and results that have been omitted due to the page limit.
Section A presents in-depth analysis of LAMDA, including detailed component analysis (\Sec{component}), computational complexity (\Sec{complexity}), LAMDA with various UDA methods (\Sec{da}), t-SNE visualization (\Sec{supp_tsne}), and the component analysis of our sampling strategy (\Sec{proto_abla}).
Section B demonstrates more results, including comparison with SSDA methods (\Sec{ssda}), comparison with cosine classifier (\Sec{supp_cosine}), and more results of DomainNet (\Sec{domainnet}).
Finally, the configuration of our DANN~\cite{dann} (\Sec{dann_config}) and implementation details of ADA methods (\Sec{ada_config}) are provided in Section C.p

\section{Further analysis of LAMDA} \label{sec:supp_analysis}
\subsection{Component analysis for each adaptation scenarios} \label{sec:component}
Table~\ref{tab:component_OH}-\ref{tab:component_RSUT} quantifies the impact of each component of LAMDA,
in every domain adaptation scenario of the two datasets.
Every component in LAMDA generally improves the performance of each domain adaptation scenario.
Comparing the second and the third rows of~\Tbl{component_RSUT}, one can see the remarkable performance gain by our label distribution matching strategy since it alleviates the significant label distribution shift of OfficeHome-RSUT.
Additionally, Our random sampling baseline incorporating DANN achieves 86.7\% on VisDA-2017 and 51.3\% on DomainNet.
\begin{table}
\centering
\caption{
Component analysis of LAMDA measured by accuracy (\%) using 10\%-budget for each source-target domain pair of four domains of OfficeHome: {\bf A}rt, {\bf C}lipart, {\bf P}roduct, and {\bf R}eal.
We evaluate from ablation baseline at the last row to LAMDA at the first row by sequentially adding three components:
(i) Prototype: prototype set sampling (o/w, sampling uniformly at random);
(ii) Matching: label distribution matching (o/w, replacing $p_i$ in Eq.~(9)-(10) with uniform distribution); and
(iii) Cosine: cosine classifier (o/w, linear classifier).
The ablation baseline is equipped with DANN~\cite{dann}.
The accuracy is a mean of three runs, and the subscript denotes standard deviation.}
\renewcommand{\arraystretch}{1.1}
\scalebox{0.7}{
\begin{tabular}{ccc|ccccccccccccc}
        \toprule
        \multirow{2}{*}{\shortstack{Prototype}} & \multirow{2}{*}{Matching} & \multirow{2}{*}{Cosine} & \multicolumn{13}{c}{OfficeHome}  \\
        
        & & & A $\veryshortarrow$ C & A $\veryshortarrow$ P & A $\veryshortarrow$ R & C $\veryshortarrow$ A & C $\veryshortarrow$ P & C $\veryshortarrow$ R & P $\veryshortarrow$ A & P $\veryshortarrow$ C & P $\veryshortarrow$ R & R $\veryshortarrow$ A & R $\veryshortarrow$ C & R $\veryshortarrow$ P & Avg  \\
        
        \midrule
        
        \cmark & \cmark & \cmark & 75.4 & 88.5 & 85.9 & 73.3 & 88.7 & 83.8 & 75.2 & 75.3 & 87.1 & 80.9 & 77.8 & 91.8 & 82.0$_{\pm \text{0.1}}$ \\
                    
        \cmark & \cmark & \xmark & 74.0 & 87.8 & 85.4 & 69.4 & 85.6 & 81.7 & 71.5 & 74.2 & 86.0 & 78.1 & 76.3 & 91.8 & 80.2$_{\pm \text{0.4}}$ \\
        
        \cmark & \xmark & \xmark & 69.6 & 82.7 & 82.8 & 64.6 & 84.2 & 78.4 & 66.3 & 72.1 & 84.9 & 74.8 & 74.1 & 91.3 & 77.1$_{\pm \text{0.0}}$ \\            
        
        \xmark & \xmark & \xmark & 65.2 & 77.4 & 79.1 & 61.3 & 77.5 & 73.9 & 65.6 & 68.0 & 81.2 & 73.7 & 69.4 & 84.6 & 73.1$_{\pm \text{0.2}}$ \\
        \bottomrule
\end{tabular}
}
\label{tab:component_OH}
\end{table}
\begin{table}
\centering
\caption{Component analysis of LAMDA measured by accuracy (\%) using 10\%-budget for each source-target domain pair of three domains of OfficeHome-RSUT: {\bf C}lipart, {\bf P}roduct, and {\bf R}eal.
We evaluate from ablation baseline at the last row to LAMDA at the first row by sequentially adding three components:
(i) Prototype: prototype set sampling (o/w, sampling uniformly at random);
(ii) Matching: label distribution matching (o/w, replacing $p_i$ in Eq.~(9)-(10) with uniform distribution); and
(iii) Cosine: cosine classifier (o/w, linear classifier).
The ablation baseline is equipped with DANN~\cite{dann}.
The accuracy is a mean of three runs, and the subscript denotes standard deviation.}
\renewcommand{\arraystretch}{1.1}
\scalebox{0.9}{
\begin{tabular}{ccc|ccccccc}
        \toprule
        \multirow{2}{*}{\shortstack{Prototype}} & \multirow{2}{*}{Matching} & \multirow{2}{*}{Cosine} & \multicolumn{7}{c}{OfficeHome-RSUT}  \\
        
        & & & C $\veryshortarrow$ P & C $\veryshortarrow$ R & P $\veryshortarrow$ C & P $\veryshortarrow$ R & R $\veryshortarrow$ C & R $\veryshortarrow$ P & Avg  \\
        
        \midrule
        
        \cmark & \cmark & \cmark & 82.0 & 74.9 & 60.8 & 81.5 & 65.4 & 85.9 & 75.1$_{\pm \text{0.8}}$ \\
                    
        \cmark & \cmark & \xmark & 79.2 & 72.6 & 61.3 & 81.0 & 62.7 & 86.2 & 73.8$_{\pm \text{0.4}}$ \\
        
        \cmark & \xmark & \xmark & 70.5 & 64.6 & 52.8 & 75.7 & 53.8 & 79.6 & 66.2$_{\pm \text{1.0}}$ \\            
        
        \xmark & \xmark & \xmark & 64.2 & 60.1 & 52.0 & 73.7 & 54.6 & 73.6 & 63.1$_{\pm \text{0.9}}$ \\
        \bottomrule
\end{tabular}
}
\label{tab:component_RSUT}
\end{table}

\begin{table}
\centering
\caption{Computational complexity and computation time (seconds and minutes) per round on two source-target domain pairs: Art to Clipart of OfficeHome and Clipart to Sketch of DomainNet,
where $n_\textT$ is the size of the target dataset, $t$ is the number of clustering iterations, $B$ is the budget, $d$ is the dimension of embedding, and $n_\textP$ is the size of the prototype set.
The query time is measured at the first round when 2\% of the target dataset is actively sampled.
($\dagger$): Sampling of $\text{S}^\text{3}$VAADA\cite{s3vaada} requires a large computation cost of network forward and backward pass proportional to $n_\textT$, since it adversarially perturbs each sample to measure uncertainty.}
\renewcommand{\arraystretch}{1.1}
\scalebox{0.82}{
\begin{tabular}{p{1cm}lccc}
        \toprule
        & \multirow{2}{*}{\shortstack{AL strategy}} & \multirow{2}{*}{\shortstack{Computational complexity}} & \multicolumn{2}{c}{Computation time} \\
        
        & & & $\;$OfficeHome (A $\veryshortarrow$ C)$\;$ & $\;$DomainNet (C $\veryshortarrow$ S)$\;$ \\
        
        \midrule
        
        \multirow{4}{*}{\rotatebox{90}{\shortstack{forward +\\sampling}}}
        & TQS\cite{tqs}                          & $O\big(n_\textT \text{log}n_\textT\big)$ & 5.41s & 0.88m \\
                    
        & CLUE\cite{clue}                        & $O\big(tn_\textT Bd\big)$ & 35.98s & 43.94m \\
        
        & $\text{S}^\text{3}$VAADA\cite{s3vaada} & $O\big(n_\textT^2(d + B^2)\big)^{\dagger}$ & 132.20s & 487.11m \\
        
        & Ours                                   & $O\big(n_\textT^2(d+n_\textP^2)\big)$ & 7.11s & 2.37m \\
        \bottomrule
\end{tabular}
}
\label{tab:complexity}
\end{table}

\subsection{Computational complexity of LAMDA} \label{sec:complexity}
In~\Tbl{complexity}, we compare the computational complexity and computation time per round of LAMDA and previous ADA methods.
The computation time is measured on Intel(R) Xeon(R) Gold 6240 CPU.
While the computational complexity of LAMDA exhibits quadratic growth regarding the target dataset's size, it is significantly faster than CLUE~\cite{clue} and $\text{S}^\text{3}$VAADA~\cite{s3vaada}, even on the large scale Clipart dataset of DomainNet consists of 30K images.
This is because the calculation for our sampling consists of simple matrix multiplication, which can be computed efficiently in parallel.
LAMDA achieves the best performance among the previous arts with a reasonable sampling time.

\subsection{Combine LAMDA with other DA methods} \label{sec:da}
\begin{figure*}[!t]
\begin{center}
\includegraphics[width=0.99 \linewidth]{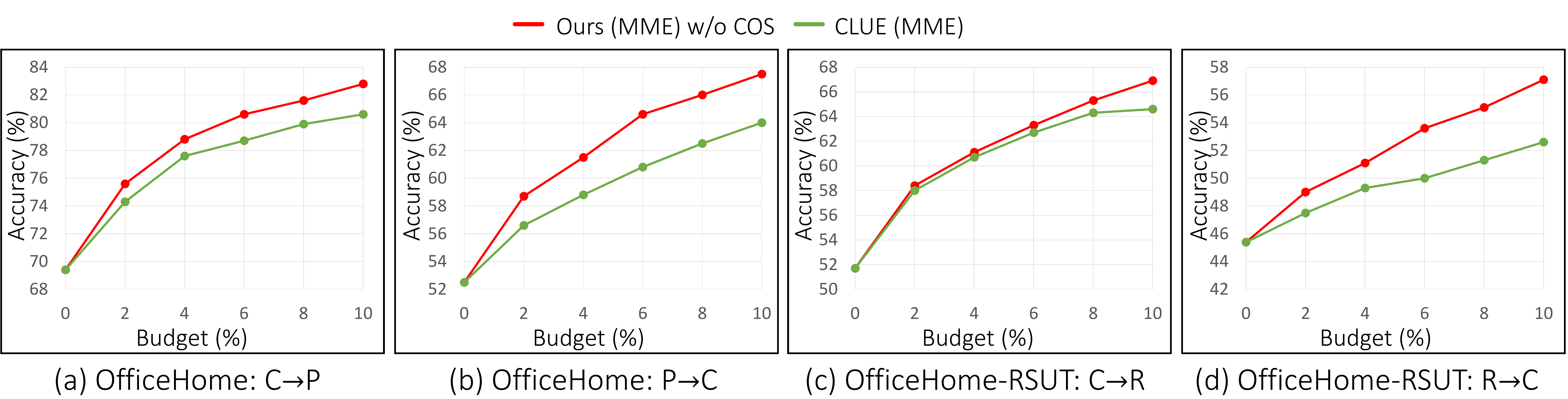}
\end{center}
\vspace{-4mm}
\caption{Accuracy versus the percent of labeled target instances as budget, when each method is equipped with MME~\cite{mme}.
The accuracies are measured on the source-target domain pair of two domains of OfficeHome: {\bf C}lipart and {\bf P}roduct; and OfficeHome-RSUT: {\bf C}lipart and {\bf R}eal).
w$\!$/$\!$o COS: Ours without cosine classifier.}
\label{fig:mme}
\end{figure*}

In~\Fig{mme}, we compare the performance of LAMDA with with CLUE~\cite{clue} varying budget, when equipped with MME~\cite{mme}, on source-target domain pair of OfficeHome: {\bf C}lipart and {\bf P}roduct; and OfficeHome-RSUT: {\bf C}lipart and {\bf R}eal.
We follow the training protocol of CLUE, where at each round, the model is provided with the budget and is trained for 20 epochs with MME loss.
While CLUE is specialized with MME, the result shows that LAMDA constantly outperforms CLUE in all four domain adaptation scenarios. 
The performance gap between LAMDA and CLUE increases as the budget increases.
In~\Tbl{mcc_oh}, we compare the performance of ADA methods coupled with MCC~\cite{jin2020minimum}, the state-of-the art UDA method, when using 10\%-budget; the cosine classifier is not used in this comparison to verify the generalization of our sampling strategy.
While MCC increases the overall performance of ADA methods, LAMDA still shows the best performance among them.
This result demonstrates that LAMDA can significantly improve the performance regardless of adaptation technique by using the budget effectively for label distribution matching and supervised learning. 
\begin{table}
\centering
\caption{
Accuracy (\%) of ADA methods coupled with MMC~\cite{jin2020minimum}
on {OfficeHome}
using 10\%-budget 
for each source-target pair
of four domains: {\bf A}rt, {\bf C}lipart, {\bf P}roduct, and {\bf R}eal.
w$\!$/$\!$o COS: Ours without cosine classifier
}
\renewcommand{\arraystretch}{1.1}
\scalebox{0.7}{
\begin{tabular}{cc|ccccccccccccc}
        \toprule
        \multirow{2}{*}{DA method} & \multirow{2}{*}{AL method} & \multicolumn{13}{c}{OfficeHome}  \\
        
         & & A $\veryshortarrow$ C & A $\veryshortarrow$ P & A $\veryshortarrow$ R & C $\veryshortarrow$ A & C $\veryshortarrow$ P & C $\veryshortarrow$ R & P $\veryshortarrow$ A & P $\veryshortarrow$ C & P $\veryshortarrow$ R & R $\veryshortarrow$ A & R $\veryshortarrow$ C & R $\veryshortarrow$ P & Avg  \\
        
        \midrule
        
        \multirow{4}{*}{\shortstack{MCC~\cite{jin2020minimum}}} & TQS\cite{tqs} & 70.9 & \textbf{91.4} & 87.8 & 74.2 & 90.1 & 84.6 & 74.4 & 72.8 & \textbf{87.3} & 78.7 & 75.6 & 92.7 & 81.7 \\
                    
        & CLUE\cite{clue} & \textbf{77.3} & 91.1 & 87.7 & 74.3 & 90.6 & 85.4 & 74.7 & 77.2 & 87.2 & 81.3 & 78.4 & 93.1 & 83.2 \\
        
        & $\text{S}^\text{3}$VAADA\cite{s3vaada} & 71.3 & 89.1 & 87.7 & 74.2 & 87.5 & 85.5 & 74.9 & 71.8 & 86.3 & 81.3 & 76.2 & 92.4 & 81.5 \\            
        
        & Ours w$\!$/$\!$o COS & 76.9 & 90.2 & \textbf{89.0} & \textbf{76.7} & \textbf{91.5} & \textbf{86.8} & \textbf{77.8} & \textbf{77.4} & 88.5 & \textbf{83.5} & \textbf{78.5} & \textbf{93.5} & \textbf{84.2} \\
        \bottomrule
\end{tabular}
}
\label{tab:mcc_oh}
\end{table}

\subsection{t-SNE visualization} \label{sec:supp_tsne}
\begin{figure*}[!t]
\begin{center}
\includegraphics[width=0.98\linewidth]{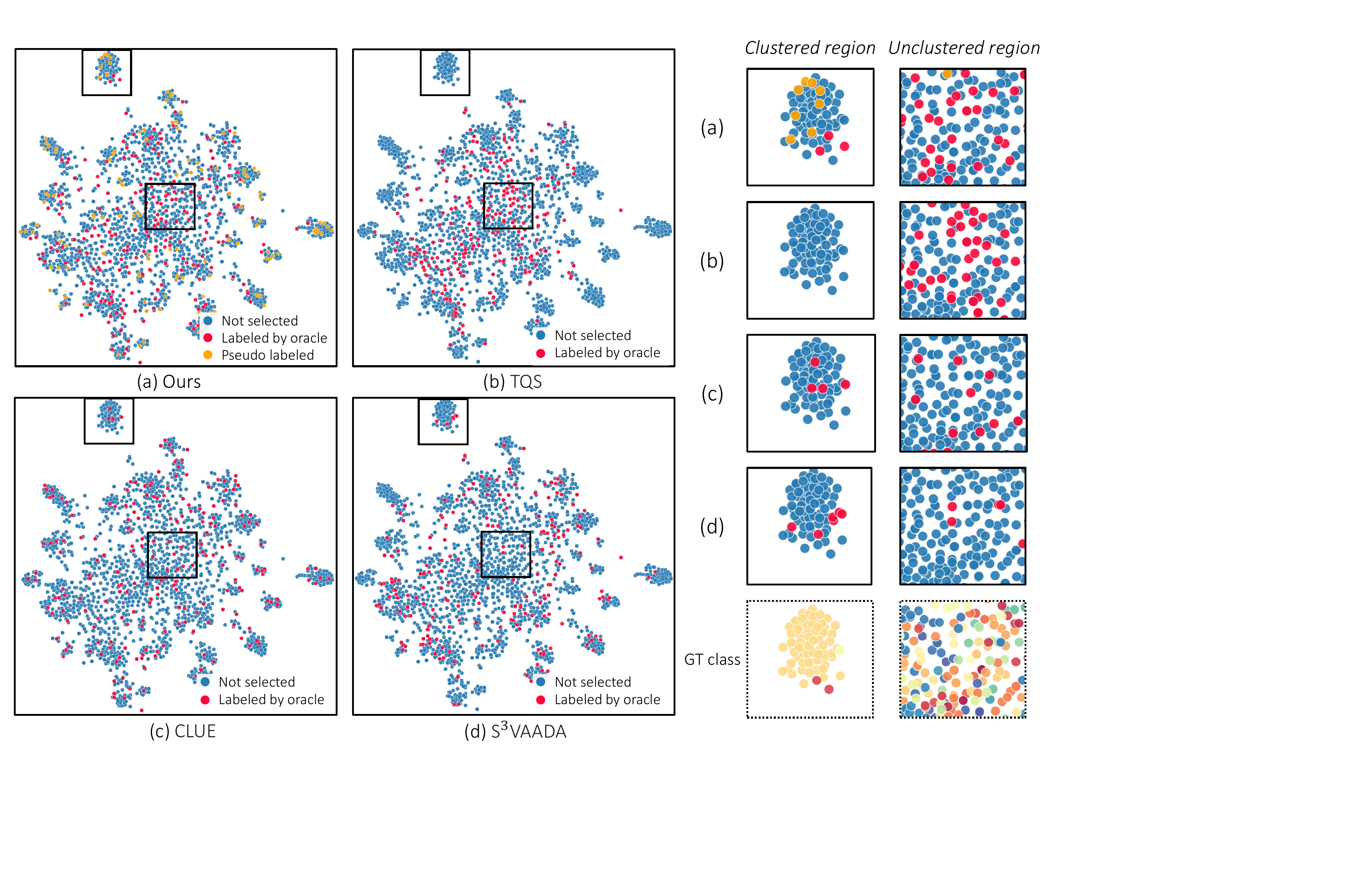}
\end{center}
\vspace{-5mm}
\caption{$t$-SNE~\cite{tsne} visualization of target feature vectors from source pre-trained model on OfficeHome Real to Art scenario.
The feature vector of the selected samples from each method are colored red.
GT class: target feature vectors colored by ground-truth class labels.}
\label{fig:supp_tsne}
\end{figure*}

In~\Fig{supp_tsne}, we visualize the target feature vectors selected by LAMDA and the previous ADA methods from the source pre-trained model.
When the feature vectors are clustered, LAMDA selects a comparably large number of prototypes within the cluster to reflect its density (\Fig{supp_tsne}a),
which provides a better estimation of target statistics as in Fig.~4.
In the clustered region, most prototypes are pseudo labeled since there tend to include easy samples (GT~class in \Fig{supp_tsne}).
While for the unclustered region, the prototypes are labeled by an oracle (\Fig{supp_tsne}a), providing ground-truth supervision to the uncertain prototypes.
This allows us to mainly invest a budget on uncertain samples while utilizing density-aware samples to estimate the target label distribution.

On the other hand, the selected samples of the previous methods less preserve the density of the target data as they avoid selecting easy samples within the clustered regions.
Even though CLUE~\cite{clue} and $\text{S}^\text{3}$VAADA~\cite{s3vaada} select part of the samples within the clustered region, they do not reflect the density of the cluster.
Since CLUE and $\text{S}^\text{3}$VAADA spend part of their budgets in the clustered regions, they cannot select samples as many as LAMDA in the unclustered regions (\Fig{supp_tsne}c and \Fig{supp_tsne}d).
Oppositely, while TQS~\cite{tqs} selects samples as many as LAMDA within the unclustered region, it ignores samples in the clustered region (\Fig{supp_tsne}b).

\subsection{Detailed component analysis of prototype set sampling} \label{sec:proto_abla}
\Tbl{analysis_proto_oh} quantifies the contribution of each components of our sampling strategy:
(i) Subset sampling using MMD measure;
(ii) Preventing labeling of the easy prototypes;
and (iii) Utilizing the pseudo-labeled prototypes for the label distribution estimation.
Every component of our sampling strategy improves the performance for every adaptation scenario in OfficeHome.
The performance gap between the last (random sampling with DANN~\cite{dann}) and the second last rows verifies that our distribution-aware sampling boosts the effect of supervised learning.
Comparing the second and the third rows, one can see the performance gain by considering uncertainty for the sampling.
The performance gap between the first and the second rows shows utilizing the pseudo-labeled prototypes improves the quality of the label distribution estimation. 
\begin{table}
\centering
\caption{
Component analysis of prototype set sampling measured by accuracy (\%) using 10\%-budget for each source-target domain pair of four domains of OfficeHome: {\bf A}rt, {\bf C}lipart, {\bf P}roduct, and {\bf R}eal.
We evaluate from ablation baseline at the last row to our sampling strategy at the first row by sequentially adding three components:
(i) Density: Sampling with MMD (o/w, sampling uniformly at random);
(ii) Uncertainty: only label uncertain prototypes using the top-2 margin (o/w, using $\Delta=1$ in~Algorithm~1; and
(iii) Pseudo label: utilizing the pseudo-labeled prototypes for the label distribution estimation (o/w, ignoring the pseudo labeled prototypes in Eq.~(7)).
The ablation baseline is equipped with DANN~\cite{dann}.}
\renewcommand{\arraystretch}{1.1}
\scalebox{0.7}{
\begin{tabular}{ccc|ccccccccccccc}
        \toprule
        \multirow{2}{*}{\shortstack{Density}} & \multirow{2}{*}{Uncertainty} & \multirow{2}{*}{Pseudo label} & \multicolumn{13}{c}{OfficeHome}  \\
        
        & & & A $\veryshortarrow$ C & A $\veryshortarrow$ P & A $\veryshortarrow$ R & C $\veryshortarrow$ A & C $\veryshortarrow$ P & C $\veryshortarrow$ R & P $\veryshortarrow$ A & P $\veryshortarrow$ C & P $\veryshortarrow$ R & R $\veryshortarrow$ A & R $\veryshortarrow$ C & R $\veryshortarrow$ P & Avg  \\
        
        \midrule
        
        \cmark & \cmark & \cmark & 75.4 & 88.5 & 85.9 & 73.3 & 88.7 & 83.8 & 75.2 & 75.3 & 87.1 & 80.9 & 77.8 & 91.8 & 82.0 \\
                            
        \cmark & \cmark & \xmark & 72.5 & 87.3 & 84.1 & 73.2 & 87.6 & 81.3 & 74.0 & 73.6 & 84.8 & 78.7 & 76.0 & 90.6 & 80.3 \\
        
        \cmark & \xmark & \xmark & 69.4 & 84.1 & 82.6 & 71.0 & 83.5 & 79.6 & 71.6 & 70.4 & 83.9 & 77.2 & 73.9 & 88.0 & 77.9 \\            
        
        \xmark & \xmark & \xmark & 66.1 & 76.7 & 79.2 & 62.2 & 76.1 & 73.6 & 66.1 & 70.3 & 82.1 & 72.9 & 71.3 & 83.1 & 73.3 \\
        \bottomrule
\end{tabular}
}
\label{tab:analysis_proto_oh}
\end{table}

\begin{figure*}
\begin{center}
\includegraphics[width = 0.75\textwidth]{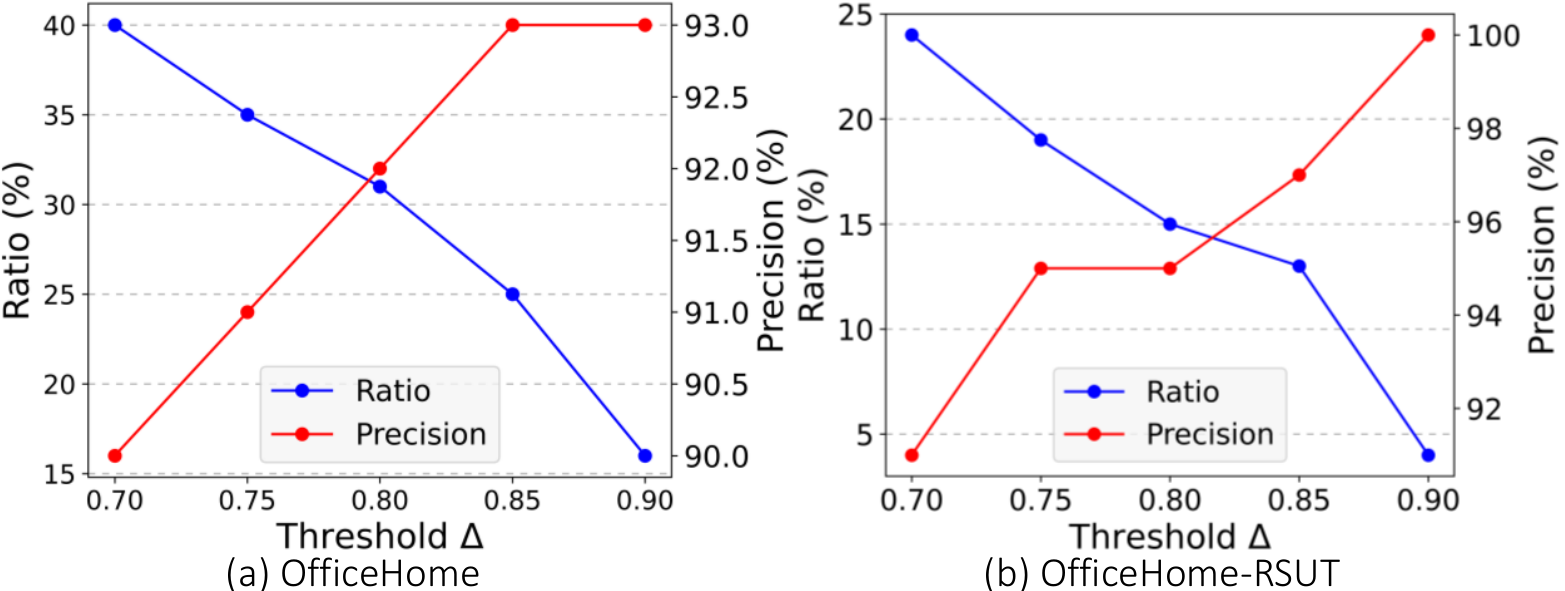}
\end{center}
\caption{The ratio (\%) of the pseudo-labeled prototypes among all the prototypes and the precision (\%) of the pseudo-labeled prototypes, along with different threshold $\Delta$.
The score is averaged over \textit{all} scenarios of OfficeHome and OfficeHome-RSUT.}
\label{fig:plbl_ratio}
\end{figure*}
\subsection{The ratio of the pseudo-labeled prototypes} \label{sec:plbl_ratio}
In~\Fig{plbl_ratio}, we measure the ratio of the pseudo-labeled prototypes among all the prototypes with different threshold values.
As shown in~\Fig{plbl_ratio}, a considerable amount of prototypes is pseudo-labeled.
When $\Delta$ increases, the portion of pseudo-labeled prototypes decreases, while their precision increases.

\section{More results} \label{sec:supp_result}
\subsection{Comparison with SOTA semi-supervised DA method} \label{sec:ssda}
Semi-Supervised Domain Adaptation (SSDA) is a label-efficient domain adaptation task that allows a small amount of labeled data per class (\ie, $k$-shot per class) in the target domain.
While SSDA and ADA are similar in that both utilize a small amount of labeled target data for domain adaptation, the focus of their methods is slightly different.
While ADA methods focus on sampling the most performance-profitable data, SSDA methods focus on effective training strategies to utilize the few labeled target data.

In Table~\ref{tab:ssda_OH}-\ref{tab:ssda_RSUT}, we compare LAMDA with state-of-the-art SSDA methods~\cite{cdac,ecacl} using a 10\%-budget for the source-target domain pair of OfficeHome and OfficeHome-RSUT.
LAMDA surpasses previous arts of both ADA and SSDA in every setting of the datasets.
The performance of state-of-the-art SSDA methods is often as competitive as or even outperforms the previous ADA methods,
Thus, combining the training strategy of SSDA with ADA methods would boost the performance of ADA.

\begin{table*} [!b]
\centering
\caption{Comparison with state-of-the-art Semi-Supervised Domain Adaptation (SSDA) methods~\cite{cdac,ecacl} measured by accuracy (\%) when using 10\%-budget for each source-target domain pair of four domains of OfficeHome: {\bf A}rt, {\bf C}lipart, {\bf P}roduct, and {\bf R}eal.
(${\dagger}$): Since ECACL requires the same number of labels for each class, we randomly sample the same number of instances within each class.}
\scalebox{0.75}{
\begin{tabular}{cc|ccccccccccccc}
        \toprule
        \multirow{2}{*}{Task} & \multirow{2}{*}{Method} & \multicolumn{13}{c}{OfficeHome}  \\
        
         & & A $\veryshortarrow$ C & A $\veryshortarrow$ P & A $\veryshortarrow$ R & C $\veryshortarrow$ A & C $\veryshortarrow$ P & C $\veryshortarrow$ R & P $\veryshortarrow$ A & P $\veryshortarrow$ C & P $\veryshortarrow$ R & R $\veryshortarrow$ A & R $\veryshortarrow$ C & R $\veryshortarrow$ P & Avg  \\
        
        \midrule
        
        \multirow{2}{*}{SSDA} & ECACL$^{\dagger}\!$~\cite{ecacl} & 72.2 & 86.7 & 82.8 & 70.5 & 85.0 & 82.6 & 70.9 & 71.5 & 82.9 & 76.0 & 74.0 & 88.9 & 78.7 \\

         & CDAC~\cite{cdac} & 69.5 & 83.2 & 80.2 & 66.9 & 82.4 & 78.7 & 66.1 & 70.6 & 80.9 & 72.3 & 70.5 & 87.2 & 75.7 \\
        
        \midrule
                    
        \multirow{4}{*}{ADA} & TQS~\cite{tqs} & 64.3 & 84.8 & 83.5 & 66.1 & 81.0 & 76.7 & 66.5 & 61.4 & 82.0 & 73.7 & 65.9 & 88.5 & 74.5 \\

         & CLUE~\cite{clue} & 62.1 & 80.6 & 73.9 & 55.2 & 76.4 & 75.4 & 53.9 & 62.1 & 80.7 & 67.5 & 63.0 & 88.1 & 69.9 \\

         & $\text{S}^\text{3}$VAADA~\cite{s3vaada} & 67.8 & 83.9 & 82.9 & 67.0 & 81.4 & 79.5 & 65.8 & 65.9 & 82.4 & 74.8 & 68.6 & 87.9 & 75.7 \\

         & Ours & \textbf{74.8} & \textbf{88.5} & \textbf{86.9} & \textbf{73.8} & \textbf{88.2} & \textbf{83.3} & \textbf{74.6} & \textbf{75.5} & \textbf{86.9} & \textbf{80.8} & \textbf{77.8} & \textbf{91.7} & \textbf{81.9} \\
        \bottomrule
\end{tabular}
}
    \label{tab:ssda_OH}
\end{table*}

\begin{table*} [!b]
\centering
\caption{Comparison with state-of-the-art Semi-Supervised Domain Adaptation (SSDA) methods~\cite{cdac,ecacl} measured by accuracy (\%) when using 10\%-budget for each source-target domain pair of three domains of OfficeHome-RSUT: {\bf C}lipart, {\bf P}roduct, and {\bf R}eal.
(${\dagger}$): Since ECACL requires the same number of labels for each class, we randomly sample the same number of instances within each class.}
\scalebox{0.85}{
\begin{tabular}{cc|ccccccccccccc}
        \toprule
        \multirow{2}{*}{Task} & \multirow{2}{*}{Method} & \multicolumn{7}{c}{OfficeHome-RSUT}  \\
        
         & & C $\veryshortarrow$ P & C $\veryshortarrow$ R & P $\veryshortarrow$ C & P $\veryshortarrow$ R & R $\veryshortarrow$ C & R $\veryshortarrow$ P & Avg  \\
        
        \midrule
        
        \multirow{2}{*}{SSDA} & ECACL$^{\dagger}\!$~\cite{ecacl} & 78.6 & 68.6 & 59.5 & 77.1 & 61.9 & 82.0 & 71.3 \\

         & CDAC~\cite{cdac} & 73.0 & 58.7 & 55.8 & 73.3 & 50.3 & 77.3 & 64.7 \\
        
        \midrule
                    
        \multirow{4}{*}{ADA} & TQS~\cite{tqs} & 69.4 & 65.7 & 53.0 & 76.3 & 53.1 & 81.1 & 66.4 \\

         & CLUE~\cite{clue} & 69.7 & 65.9 & 57.1 & 73.4 & 59.5 & 82.7 & 68.1 \\

         & $\text{S}^\text{3}$VAADA~\cite{s3vaada} & 73.0 & 63.0 & 50.7 & 69.6 & 52.6 & 78.3 & 64.5 \\

         & Ours & \textbf{81.2} & \textbf{75.7} & \textbf{64.1} & \textbf{81.6} & \textbf{65.1} & \textbf{87.2} & \textbf{75.8} \\
        \bottomrule
\end{tabular}
}
\label{tab:ssda_RSUT}
\end{table*}

\subsection{Previous methods equipped with cosine classifier} \label{sec:supp_cosine}
\begin{figure*}[!t]
\begin{center}
\includegraphics[width=0.78 \linewidth]{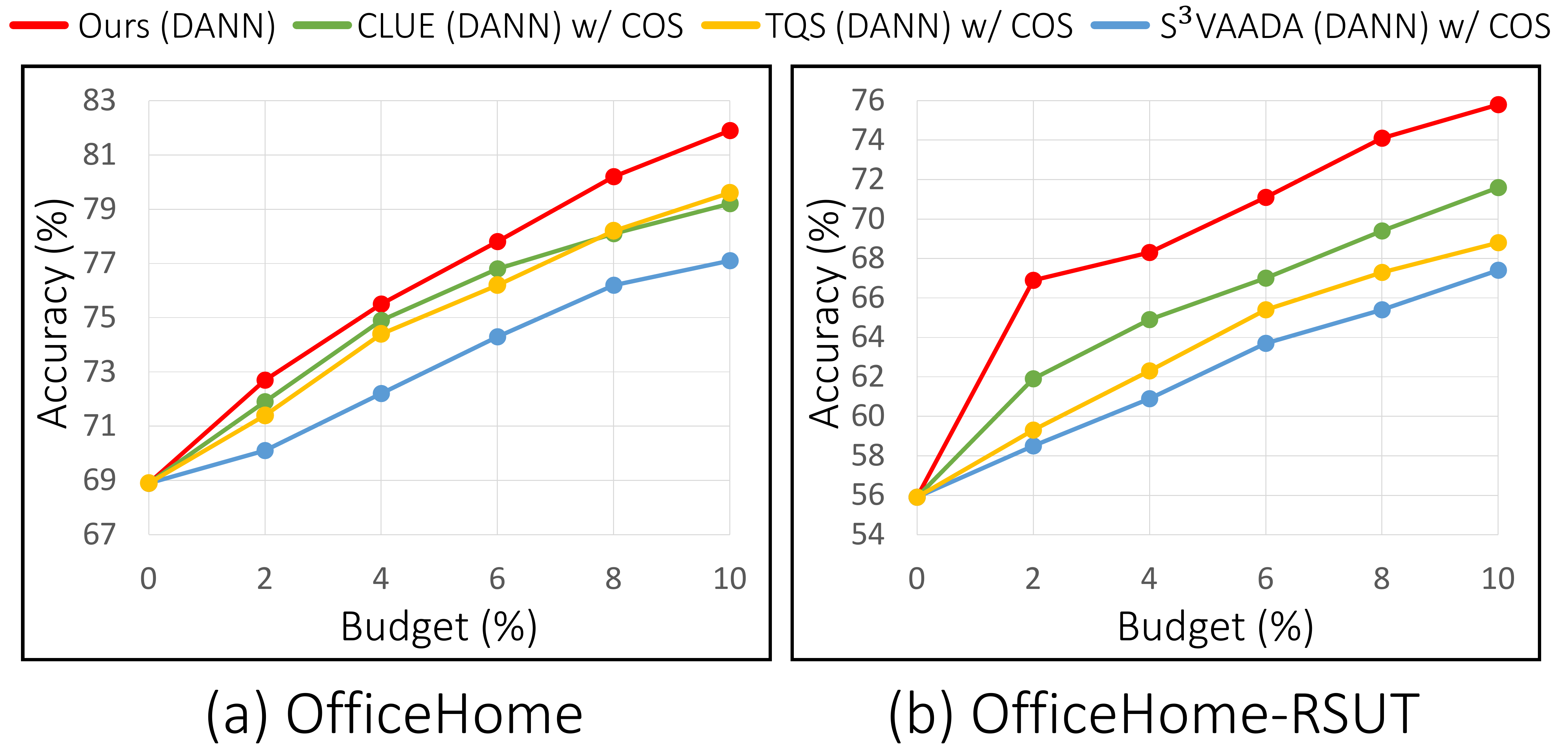}
\end{center}
\caption{Accuracy versus the percent of labeled target instances as budget, when each method is equipped with DANN~\cite{dann} and cosine classifier.
The accuracies are averaged on \textit{all} scenarios of the {OfficeHome} and {OfficeHome-RSUT}.
W/ COS: with cosine classifier.}
\label{fig:cos_acc}
\end{figure*}

In~\Fig{cos_acc}, we combine the cosine classifier with the existing approaches and compare their performance with that of LAMDA varying budget for each of OfficeHome and OfficeHome-RSUT datasets.
hods with 10\%-budget.
LAMDA constantly outperforms the previous arts in every setting on both datasets, where it with 6\%-budget is often as competitive as or even surpasses the previous methods with 10\%-budget.
The performance gap between LAMDA and other methods increases as the budget increases (\Fig{cos_acc}a), indicating that LAMDA effectively utilizes the budget in label distribution matching and supervised learning.

\subsection{Evaluation on DomainNet} \label{sec:domainnet}
\begin{figure*}[!t]
\begin{center}
\includegraphics[width=0.75 \linewidth]{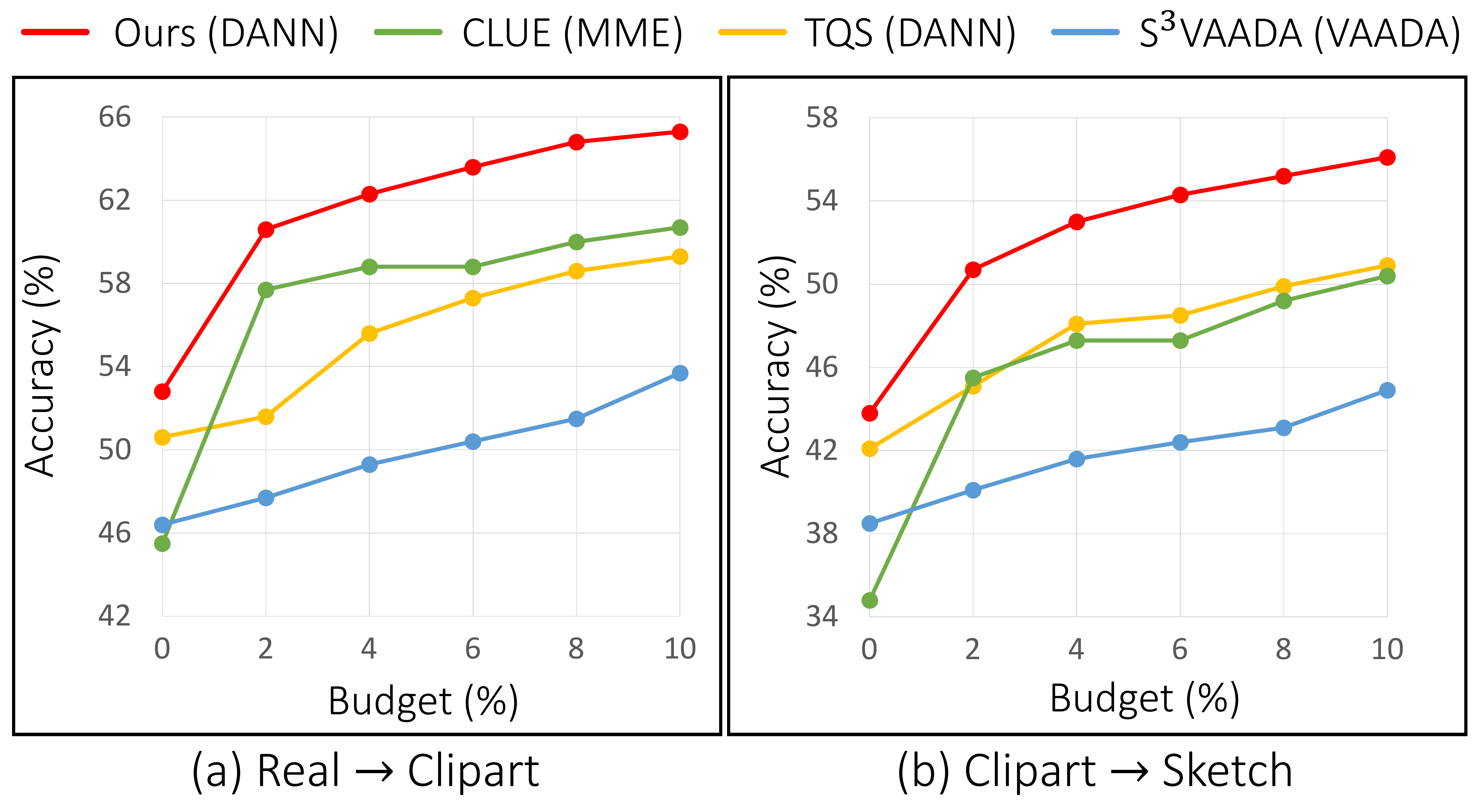}
\end{center}
\caption{Accuracy versus the percent of labeled target instances as budget for two source-target domain pair of DomainNet: (a) Real to Clipart, (b) Clipart to Sketch.}
\label{fig:domainnet}
\end{figure*}

In~\Fig{domainnet}, we compare the performance of LAMDA and the existing methods varying budget for two source-target domain pairs: Art to Product and Product to Clipart of DomainNet.
LAMDA clearly outperforms the previous arts for the two domain adaptation settings, which demonstrates the scalability of LAMDA in the large-scale datasets.
In particular, LAMDA with only 2\%-budget is often as competitive as or even outperforms the previous methods with a 10\%-budget.

\section{Experiment details} \label{sec:details}

\subsection{Configuration of DANN}  \label{sec:dann_config}
We adopt the implementation of DANN~\cite{dann} from~\cite{dalib} and follow its default configurations.
The discriminator consists of three fully connected layers of $1024$, where each hidden layer is followed by batch normalization layer~\cite{Batchnorm} and ReLU.
We utilize Gradient Reverse Layer (GRL) for adversarial learning, where the coefficient $\lambda$ of GRL is scheduled with $\lambda(s) = \frac{2}{1+\text{exp}(-s)}$, where $s$ denotes the training progress that scales from 0 to 1. 
We use the same DANN configuration and training schedule when combining previous ADA methods with DANN.

\subsection{Implementation details of previous ADA methods} \label{sec:ada_config}
We evaluate the previous ADA methods based on the official implementations of TQS\footnote{https://github.com/thuml/Transferable-Query-Selection}~\cite{tqs},  CLUE\footnote{https://github.com/virajprabhu/CLUE}~\cite{clue}, and $\text{S}^\text{3}$VAADA\footnote{https://github.com/val-iisc/s3vaada}~\cite{s3vaada}.
We fit their implementations into our evaluation protocol with minimal modifications.
We change the architecture of CLUE from ResNet34 into ResNet50, and we add missing parts of TQS implementation to make the code follow the original paper.
When training each method, we use hyper-parameter values stated in each original paper, and if not stated, we tune them using the validation set.

\clearpage

\end{document}